\documentclass[preprint,12pt]{elsarticle}
\usepackage{amssymb}
\usepackage{multirow}
\usepackage{threeparttable}
\usepackage{xcolor}
\usepackage{soul}
\sethlcolor{yellow}
\usepackage{longtable}
\usepackage{booktabs}
\usepackage{makecell}
\usepackage{url}
\usepackage{adjustbox}

\usepackage{caption}
\newcolumntype{Y}{>{\centering\arraybackslash}X}
\usepackage{algorithm,algcompatible,algpseudocode}
\usepackage{graphicx} 
\usepackage{tabularx, array}
\newcolumntype{Y}{>{\raggedright\arraybackslash}X}
\usepackage{amsmath}
\usepackage{amsthm}
\usepackage{pifont}

\begin{document}

\begin{frontmatter}

\title{Self-Supervised Multi-Scale Transformer with Attention-Guided Fusion for Efficient Crack Detection}

%% Authors (all affiliated with North Dakota State University)
\author[1]{Blessing Agyei Kyem}
\ead{blessing.agyeikyem@ndsu.edu}

\author[1]{Joshua Kofi Asamoah}
\ead{joshua.asamoah@ndsu.edu}

\author[1]{Eugene Denteh}
\ead{eugene.denteh@ndsu.edu}

\author[1]{Andrews Danyo}
\ead{andrews.danyo@ndsu.edu}

\author[1]{Armstrong Aboah\corref{cor1}}
\ead{armstrong.aboah@ndsu.edu}

%% Corresponding author footnote
\cortext[cor1]{Corresponding author}

%% Single shared affiliation
\affiliation[1]{%
  organization={Department of Civil, Construction and Environmental Engineering, North Dakota State University},%
  % addressline={1360 Bolley Drive},%
  city={Fargo},%
  postcode={58102},%
  state={ND},%
  country={USA}%
}

%% Abstract
\begin{abstract}
Pavement crack detection has long depended on costly and time-intensive pixel-level annotations, which limit its scalability for large-scale infrastructure monitoring. To overcome this barrier, this paper examines the feasibility of achieving effective pixel-level crack segmentation entirely without manual annotations. Building on this objective, a fully self-supervised framework, Crack-Segmenter, is developed, integrating three complementary modules: the Scale-Adaptive Embedder (SAE) for robust multi-scale feature extraction, the Directional Attention Transformer (DAT) for maintaining linear crack continuity, and the Attention-Guided Fusion (AGF) module for adaptive feature integration. Through evaluations on ten public datasets, Crack-Segmenter consistently outperforms 13 state-of-the-art supervised methods across all major metrics, including mean Intersection over Union (mIoU), Dice score, XOR, and Hausdorff Distance (HD). These findings demonstrate that annotation-free crack detection is not only feasible but also superior, enabling transportation agencies and infrastructure managers to conduct scalable and cost-effective monitoring. This work advances self-supervised learning and motivates pavement cracks detection research.
\end{abstract}


% \begin{highlights}
% \end{highlights}

%% Keywords
\begin{keyword}
Self-supervised segmentation \sep Crack detection \sep Multi-scale transformer \sep Directional attention \sep Infrastructure monitoring \sep Annotation-free learning \sep Road maintenance automation
\end{keyword}

\end{frontmatter}

% \linenumbers 


\section{Introduction}
\label{intro}
Transportation infrastructure, particularly road networks, is critical for public safety and economic development. Roads connect communities, facilitate commerce, and ensure efficient movement of people and goods. However, the constant exposure of roads to traffic loads and weather conditions gradually weakens their structural integrity, often resulting in surface cracks. Even tiny cracks that develop on these roads can quickly grow into severe defects such as potholes or large pavement failures if they are not detected and repaired early. This makes preventive maintenance very vital in prolonging the lifespan of pavements. Research indicates that preventive maintenance on small pavement cracks can reduce future repair costs by approximately 50–70\%, highlighting the financial benefit of timely intervention \cite{ Kyem2024PaveCapTF, barman2019cracksealing}. Accurate, pixel-level crack detection maps are thus essential for enabling early and low-cost pavement maintenance. These detailed maps can guide agencies in prioritizing repairs, allocating budgets efficiently, and reducing overall maintenance expenses.

To generate such maps automatically, researchers have explored various supervised learning approaches that achieve strong performance in pavement crack segmentation tasks. However, these fully-supervised methods require extensive pixel-level annotations during training, which presents significant practical challenges. This annotation burden especially for large-scale pavement datasets has motivated exploration of alternative learning paradigms \cite{BlessingAIC25}.

To address these annotation challenges, researchers have developed weakly supervised and semi-supervised approaches that aim to reduce labeling requirements while maintaining segmentation accuracy. These methods typically use coarse labels, bounding boxes, or limited annotations combined with various learning strategies. While these approaches reduce annotation burden compared to fully supervised methods, they still require some form of manual supervision during training.

More recently, some studies \cite{claim_ssl, gans_ssl} have explored self-supervised learning for crack segmentation, claiming to eliminate annotation requirements entirely. However, many of these approaches still rely on ground truth data at various stages of their pipeline, such as during pretraining or pseudo-label generation, thus not achieving truly annotation-free learning. This limitation highlights the need for a robust, fully self-supervised framework that can accurately segment pavement cracks without any manual annotations during training while achieving competitive performance.

To address these limitations, the paper propose an efficient, fully self-supervised segmentation framework designed specifically for pavement crack segmentation. Unlike existing approaches, the proposed method requires no manual annotations, pixel labels, or any form of ground truth supervision, thus significantly reducing the cost and time associated with data preparation. The proposed framework introduces three key innovations: a Scale-Adaptive Embedder that captures multi-resolution crack features, a Directional Attention Transformer that preserves linear crack structures, and an Attention-Guided Fusion module that adaptively combines multi-scale representations. These components work together with novel consistency losses to enable effective self-supervised learning for crack segmentation.
The main contributions of the proposed approach are summarized below:

\begin{itemize}
\item Crack-Segmenter is introduced as an end-to-end self-supervised framework for pavement crack segmentation that eliminates the need for pixel-level or weak annotations, thereby substantially reducing annotation costs.
\item  Three modules are designed: \emph{the Scale-Adaptive Embedder} for multi-resolution feature extraction, the \emph{Directional Attention Transformer} to preserve elongated crack geometry, and the \emph{Attention-Guided Fusion} module to adaptively merge scale-specific representations.
\item Inter-scale and intra-scale consistency losses are developed to enhance coherent feature representations, enabling more effective model learning without manual supervision.
\item The framework was evaluated on ten public crack datasets against thirteen state-of-the-art fully supervised models and consistently outperformed all baselines with statistically significant gains.
\end{itemize}

\section{Related Works}
\label{lit_review}
Existing crack segmentation approaches can be categorized into three main paradigms based on their annotation requirements: fully supervised methods that rely on complete pixel-level labels, weakly and semi-supervised methods that use partial or coarse annotations, and self-supervised methods. As such, this section reviews representative methods from each category, with particular emphasis on identifying gaps in truly annotation-free crack segmentation.

\subsection{Fully-Supervised Crack Segmentation Methods}
\label{supervised_lit_review}
Fully supervised methods have utilized rich pixel-level annotations to develop specialized models for accurate pavement crack segmentation. Several high-performing approaches have demonstrated the effectiveness of full supervision. For instance, Lau et al. \cite{Lau2020Automated} replaced the U-Net encoder with a pretrained ResNet-34 and reported F1 scores of 96\% on CFD and 73\% on Crack500, proving that full-mask supervision can yield strong accuracy. Similarly, \cite{kyem_context_cracknet} proposed Context-CrackNet, a supervised model using global and local attention modules to accurately segment tiny and large pavement cracks. \cite{PDSNET} also developed PDSNet, a supervised deep learning framework for segmenting multiple asphalt pavement distresses using 2D and 3D images, achieving a high mean Intersection over Union (MIoU) of 83.7\%. However, these approaches required extensive annotations during training, with \cite{PDSNET} requiring manual annotation of 5000 images with pixel-level labels, highlighting the labor-intensive nature of fully supervised learning.

Early fully supervised approaches utilized basic architectures that were later improved through architectural innovations. Dung et al. \cite{DUNG201952} adopted a Fully Convolutional Network (FCN) based on a VGG16 encoder for automated concrete crack detection, where high-level features were upsampled via deconvolution to produce crack masks. While this approach proved the feasibility of end-to-end crack segmentation, the basic FCN architecture struggled with limited context and spatial consistency, often missing fine crack details.
 
Researchers subsequently improved crack segmentation by integrating multi-scale feature fusion and boosting techniques. Yang et al. \cite{Yang_fpn} introduced a Feature Pyramid and Hierarchical Boosting Network (FPHBN) that fuses semantic information from deep and shallow layers and utilizes a feature pyramid to enhance detection of cracks at different scales. This method improved the generalization to various crack widths and backgrounds by considering multi-level features simultaneously. Similarly, Liu et al. \cite{deepcrack} proposed a deeply supervised encoder–decoder network (DeepCrack) which aggregates multi-scale features from multiple network layers, capturing both fine and coarse crack patterns for more robust segmentation. These multi-scale approaches demonstrated higher accuracy than plain FCNs, but the heavy pooling in their backbones could still cause some loss of fine spatial information.
To better preserve crack locality, later studies adopted encoder–decoder architectures like U-Net. Jenkins et al. \cite{jenkins2018deep} showed that a vanilla U-Net can effectively segment road cracks at the pixel level, and further improvements added attention gating to suppress irrelevant background features. Building on this idea, Pan et al. \cite{PAN2020103357} developed an attention-enhanced U-Net variant called SCHNet, which incorporates parallel spatial, channel, and feature-pyramid attention modules into a VGG19-based network.

Transformer-based architectures have also been explored: Guo et al. \cite{GUO2023104646} proposed a Crack Transformer (CT) model using a Swin Transformer encoder and all-MLP decoder, showing robust performance in detecting long, complex cracks even under noisy conditions. Similarly, \cite{CrackFormer} embedded a Transformer encoder within a U-shaped CNN in their CrackFormer network, substantially improving segmentation continuity and accuracy for thin cracks. However, these supervised approaches are heavily dependent on pixel-level annotations, which are time-consuming, costly, and practically difficult to obtain at a large scale \cite{Kyem2024AdvancingPD, ASAMOAH2025128003}.

\subsection{Weakly-Supervised and Semi-Supervised Crack Segmentation}
\label{semi_weakly_lit_review}
To alleviate the burden of dense labeling, researchers have developed weakly supervised frameworks that learn from coarse annotations (e.g. image-level labels). For example, Xiang et al. \cite{XIANG2024108497} introduced UWSCS, a crack segmentation framework that uses limited coarse labels alongside superpixel and shrink-based correction modules to train a dual encoder network. Similarly, recent studies have integrated the Segment Anything Model (SAM) with interactive segmentation using bounding box prompts and deep transfer learning to enable semi supervised crack detection \cite{LI2025105899, pavesam}. However, these approaches still rely on manual bounding box annotations, which remain costly and time consuming, particularly for large scale pavement crack datasets \cite{MuturiAIC25}.

To further reduce annotation dependency, some recent frameworks have incorporated advanced learning strategies such as adversarial training \cite{Li2020Semi-Supervised, adversarial_gans, stargan}, student teacher learning \cite{Wang2021Semi-supervised, triplet_teacher, teacher_pseudo}, and graph based modeling using graph convolutional networks \cite{Feng2021GCN-Based, contrastive, Denteh2025IntegratingTB}. These techniques are often used within weakly or semi supervised architectures to enhance learning from limited annotations, but they still depend on some form of manual supervision during training \cite{Denteh2025DemographicsInformedNN}.

Beyond these general strategies, specific weakly supervised methods have been developed for crack segmentation. Al-Huda et al. \cite{Al-Huda2023} proposed a two-stage weakly supervised method based on class activation mapping and iterative refinement: a crack classification network first produces initial pixel indications of cracks at multiple scales, then a U-Net with an attention mechanism is trained on these noisy masks and incrementally fine-tunes the predictions. He et al. \cite{HE2024134668} similarly employed image-level labels to drive crack segmentation by using a generative adversarial localization strategy. They trained a U-GAT-IT model to generate class activation maps of cracks, iteratively erased detected regions to discover new crack areas, and then converted these refined maps into pseudo-labels to train a segmentation network.

Semi-supervised methods have also emerged to exploit unlabeled roadway images alongside limited labeled data, further reducing the need for annotations. Shim et al. \cite{Shim} pioneered an adversarial learning-based semi-supervised segmentation approach for concrete crack detection. Their model employed multiscale feature extractors and a generative adversarial network to train on labeled and unlabeled data simultaneously. Building on consistency-driven learning, Shi et al. \cite{SHI2025109683} developed a crack segmentation model that enforces mutual consistency constraints between dual network predictions and incorporates a boundary-aware loss. Another innovative strategy is the two-stage CrackDiffusion framework proposed by Han et al. \cite{HAN2024105332}, which combines an unsupervised anomaly detection stage with a supervised refinement stage. In the first stage, a diffusion-based inpainting model removes cracks from images to generate crack-free counterparts and uses the differences (via structural similarity measures) to localize cracks without labels. In the second stage, those initial crack maps inform a U-Net segmentation model that learns to produce precise crack masks.

Despite these advancements, weakly and semi-supervised segmentation methods still depend on pseudo-labels or sparse annotations. This requirement makes it impractical to scale to large, high-resolution pavement crack datasets that completely lack pixel-level labels, motivating the exploration of fully self-supervised approaches \cite{OworAIC25}.

\subsection{Limitations of Current Self-Supervised Crack Segmentation Approaches}
\label{fully_sup_lit_review}
Self-supervised learning (SSL) is a representation learning paradigm in which the supervisory signal is derived automatically from the structure of the unlabelled data itself, rather than from externally provided human annotations \cite{Gui2023ASO}. In the context of semantic segmentation, self-supervised segmentation extends this idea from learning image-level representations to learning pixel-level assignments without any human-drawn masks. The model first designs an intrinsic task such as grouping pixels with similar colour statistics, enforcing consistency between differently augmented views, and then treats the resulting pseudo-labels as ground truth for training \cite{Caron2018DeepCF, balestriero2023cookbookselfsupervisedlearning}. Because these labels arise automatically from the data, the pipeline needs no external annotations at any stage.

SSL has been applied to pavement and other civil-infrastructure images through tasks such as self-training for cracks \cite{BEYENE2023107889, NGUYEN2025105892}, pavement-surface anomaly detection \cite{LIN2022104544, anomaly_pave}, sidewalk-quality classification \cite{sidewalk}. However, research in the area of fully self-supervised pavement distress segmentation remains limited and underexplored. Moreover, existing pavement crack segmentation techniques that claim to be fully self-supervised still suffer from significant limitations.

Although several recent studies describe their approaches as fully self supervised, many of them still rely on ground truth annotations at some point in the pipeline, such as during pretraining, pseudo label generation, or model calibration \cite{claim_ssl, gans_ssl}. For example, \cite{claim_ssl} propose SS-YOLO, a YOLOv8-based crack segmentation model that fuses CBAM and Gaussian multi-head self-attention with curriculum learning–driven pseudo-labeling. However, SS-YOLO still starts from a fully supervised YOLOv8 backbone trained on real crack masks before it ever "self-labels" unannotated data, so it isn't end-to-end self-supervised.

Similarly, Zhang et al. \cite{gans_ssl} proposed a dual cycle-GAN that learns to translate crack image patches into GT-like structure patterns (and back) using an unpaired "structure library" of binary skeletons. However, because that library is built from pixel-precise, human-annotated curves (e.g., VOC object‐boundary masks, Berkeley contour annotations, and public crack GTs), the method still depends on existing pixel-level labels and isn't truly end-to-end self-supervised.

Song et al. \cite{two_stage_unet} also proposed a two-stage pavement-crack framework: an improved U-Net (U-Net augmented with residual blocks and attention gates) is first contrastively pre-trained on unlabeled crack (background) patches, then fully fine-tuned with pixel-level ground-truth masks during the second stage. Because the network does not directly learn the image-to-mask mapping until this second stage where every gradient is computed against human-annotated labels, the approach is merely semi-supervised, not an end-to-end fully self-supervised segmentation method.

Ma et al. \cite{Ma2024UPCrackNetUP} introduced UP-CrackNet, a pavement-crack detector that trains a conditional GAN to inpaint randomly masked regions of crack-free road images. At inference, cracks are segmented by thresholding the pixel-wise reconstruction residuals, allowing annotation-free training. One problem with UP-CrackNet is that since it never sees real crack patterns during optimization, it must treat every unfamiliar texture as a defect. This makes the network yield a high error map and treats label artifacts present in the crack image (e.g. leaf, tyre mark) as a crack, inflating false positives that a crack-aware model could reject.

These limitations demonstrate that existing approaches fail to achieve truly annotation-free crack segmentation, highlighting the need for an end-to-end self-supervised framework that can learn effective crack representations without any manual supervision throughout the entire pipeline. 

\section{Methodology}
This section presents the proposed fully self-supervised crack segmentation framework and its architectural components. The methodology begins by formally defining the crack segmentation problem and establishing the learning objectives for annotation-free training. Subsequently, detailed descriptions of the three core modules (Scale Adaptive Embedder, Directional Attention Transformer, and Attention-Guided Fusion) are provided, followed by the loss functions used.

\subsection{Problem Structure and Overview}

Pavement binary crack segmentation involves accurately classifying every pixel in an image as either crack or non-crack. Currently, this task relies heavily on extensive ground truth pixel-level annotations (masks). That is, given a batch of input pavement images \(I \in \mathbb{R}^{B \times H \times W \times C}\) where \( B \), \( H \), \( W \), and \( C \) denote the batch size, height, width, and number of channels respectively, the goal of this task is to predict their corresponding binary segmentation maps \(S \in \{0,1\}^{B \times H \times W \times 1}\), where each pixel is assigned a value of 1 if it belongs to a crack and 0 otherwise. This supervised learning approach has achieved significant success in binary crack segmentation. However, obtaining the precise ground truth annotations for supervised learning is both costly and impractical for large-scale crack datasets.

To address this limitation, the paper propose a fully self-supervised framework that completely eliminates the need for ground truth labels or masks by learning robust feature representations directly from the input images. The proposed framework utilizes multi-scale feature extraction, directional attention mechanisms, and adaptive scale fusion to identify and segment pavement cracks accurately. The challenge here is formulating a learning paradigm that can effectively differentiate crack pixels from non-crack pixels without explicit supervision, thus making it scalable and cost-efficient.

\subsection{Overall Framework}

\begin{figure}[H]
    \centering
    \includegraphics[width=1.0\textwidth]{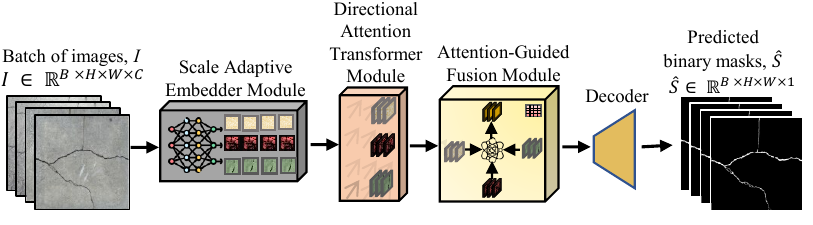}  % change file name and width as needed
    \caption{Overall framework for the proposed architecture}
    \label{fig:overall framework}
\end{figure}

The proposed self-supervised segmentation framework comprises of three primary modules designed to collectively address the aforementioned challenge: the \textbf{Scale Adaptive Embedder} (\(\Phi_{\mathrm{SAE}}\)), \textbf{Directional Attention Transformer} (\(f_{\mathrm{DAT}}\)), and \textbf{Attention-Guided Fusion} (\(f_{\mathrm{AGF}}\)). The overall framework for the proposed architecture has been shown in Figure \ref{fig:overall framework}.

The proposed framework begins with the Scale Adaptive Embedder module, which processes a batch of input images \(I\) simultaneously at multiple scales: fine, small, and large. Specifically, given a batch of input image \(I\), the module produces three distinct feature embeddings with embedding dimension \(D\):
\begin{equation}
\Phi_{\mathrm{SAE}}(I) = \{F_{f}, F_{s}, F_{l}\},
\end{equation}
where \(F_{f}, F_{s} \in \mathbb{R}^{B \times D \times H \times W}\) and \(F_{l} \in \mathbb{R}^{B \times D \times \tfrac{H}{2} \times \tfrac{W}{2}}\). \(F_{f}, F_{s}, F_{l}\) correspond to the fine, small, and large scale feature embeddings respectively. This multi-scale representation ensures comprehensive capture of cracks of varying widths and complexities, enhancing sensitivity to fine details and broad spatial context simultaneously.

Next, each scale-specific feature embedding from the Scale Adaptive Embedder undergoes further refinement through the Directional Attention Transformer module. This transformer applies efficient attention with directed convolutions to maintain and emphasize linear crack structures within feature maps, essential for accurate pavement crack identification. Formally, this transformation is represented as:
\begin{equation}
F'_{f} = f_{\mathrm{DAT}}(F_{f})
\end{equation}
\begin{equation}
F'_{s} = f_{\mathrm{DAT}}(F_{s})
\end{equation}
\begin{equation}
F'_{l} = f_{\mathrm{DAT}}(F_{l})
\end{equation}

where \(F'_{f}, F'_{s} \in \mathbb{R}^{B \times D \times H \times W}\) and \(F'_{l} \in \mathbb{R}^{B \times D \times \tfrac{H}{2} \times \tfrac{W}{2}}\). \(F'_{f}\), \(F'_{s}\), \(F'_{l}\) represent the refined feature embeddings for the different scales after passing through the Directional Attention Transformer module. This module enhances local contextual consistency within each scale, preserving crucial spatial relationships pertinent to cracks.

Finally, the Attention-Guided Fusion module integrates these refined scale-specific features from the Directional Attention Transformer module using attention-based adaptive weighting to form a unified, robust representation. Specifically, the large-scale feature map \(F'_{l}\) is upsampled and projected to match the spatial dimensions of the other scales. Subsequently, the fusion module computes attention weights and merges the scales:
\begin{equation}
F_{\mathrm{fused}} = f_{\mathrm{AGF}}\bigl(F'_{f},\, F'_{s},\, \mathrm{upsample}(F'_{l})\bigr),
\end{equation}
where \(F_{\mathrm{fused}} \in \mathbb{R}^{D \times H \times W}\) and \(upsample\) represents the upsampling operation. This adaptive fusion ensures optimal integration of both detailed crack structures and broader contextual information.

Finally, the fused feature representation \(F_{\mathrm{fused}}\) is processed through a linear decoding layer to produce the final segmentation map prediction \(\hat{S}\):
\begin{equation}
\hat{S} = f_{\mathrm{decode}}(F_{\mathrm{fused}}),
\end{equation}
where \(f_{\mathrm{decode}}\) is the decoding layer and \(\hat{S} \in \mathbb{R}^{B \times H \times W \times 1}\) represents the predicted segmentation maps.

Through cross-scale consistency losses, specifically inter-scale and intra-scale self-supervised losses, the model learns to produce consistent and accurate segmentation predictions without any ground truth annotations (masks). Thus, the proposed framework efficiently addresses pavement crack segmentation challenges, significantly reducing annotation costs while maintaining high segmentation performance. The next section goes into details on the modules used. 

\subsection{Scale Adaptive Embedder}
\begin{figure}[H]
    \centering
    \includegraphics[width=1.0\textwidth]{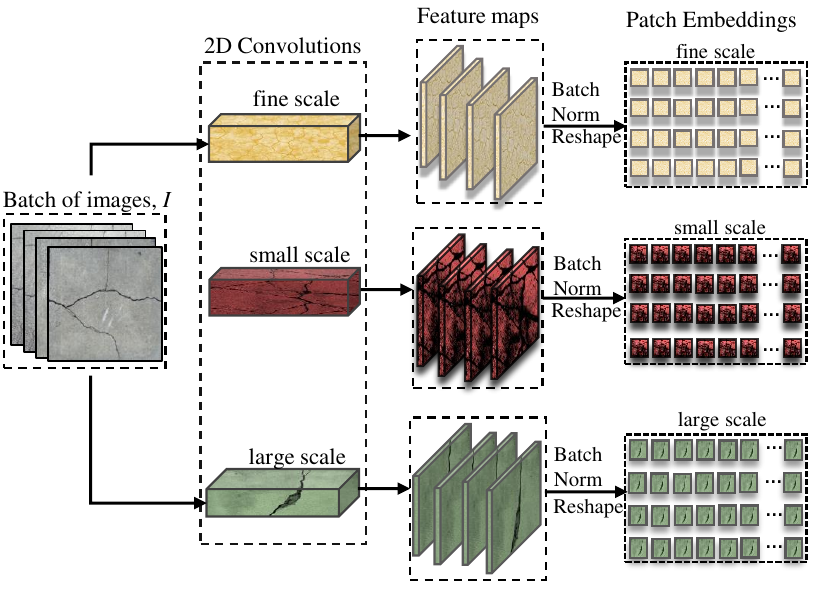}  % change file name and width as needed
    \caption{Scale Adaptive Embedder Module.}
    \label{fig:SAE module}
\end{figure}

The \textbf{Scale Adaptive Embedder (SAE)} module is designed to effectively capture multi-scale spatial information from pavement images, crucial for accurately identifying cracks of varying sizes and complexities. This module embeds the input image into feature representations at fine, small, and large spatial scales, enhancing the model’s ability to detect both detailed and broad crack structures simultaneously. Figure \ref{fig:SAE module} shows the architecture of the Scale Adaptive Embedder. 

Given a batch of input pavement images \(I \in \mathbb{R}^{B \times C \times H \times W}\), the SAE module first applies convolutional projection operations at three distinct scales to produce feature maps tailored for fine, small, and large-scale analyses.

For the fine-scale embedding, the convolution operation is mathematically defined as:
\begin{equation}
F_{f} = \sigma\bigl(W_{f} \ast I + b_{f}\bigr), \quad F_{f} \in \mathbb{R}^{B \times D \times H \times W},
\end{equation}
where \(W_{f} \in \mathbb{R}^{D \times C \times 1 \times 1}\) represents the convolutional kernel weights, \(b_{f} \in \mathbb{R}^{D}\) are biases, \(\ast\) denotes the convolution operation, and \(\sigma\) is a nonlinear activation function. This fine-scale operation ensures the capture of detailed and fine-grained crack patterns.

Similarly, for the small-scale embedding, the convolution operation is defined as:
\begin{equation}
F_{s} = \sigma\bigl(W_{s} \ast I + b_{s}\bigr), \quad F_{s} \in \mathbb{R}^{B \times D \times H \times W},
\end{equation}
where \(W_{s} \in \mathbb{R}^{D \times C \times 3 \times 3}\) with appropriate padding and stride set to 1 to maintain spatial dimensions. This operation captures crack structures at intermediate spatial resolutions, preserving local contextual relationships.

The large-scale embedding convolution operation is expressed as:
\begin{equation}
F_{l} = \sigma\bigl(W_{l} \ast I + b_{l}\bigr), \quad F_{l} \in \mathbb{R}^{B \times D \times \tfrac{H}{2} \times \tfrac{W}{2}},
\end{equation}
where \(W_{l} \in \mathbb{R}^{D \times C \times 3 \times 3}\) with stride set to 2 and appropriate padding to reduce spatial dimensions. This large-scale embedding helps in capturing broader spatial contexts and large-scale pavement defects.

After convolutional projections, each feature map undergoes batch normalization to stabilize learning and enhance convergence:
\begin{equation}
F_{f}' = \mathrm{BatchNorm}(F_{f})
\end{equation}
\begin{equation}
F_{s}' = \mathrm{BatchNorm}(F_{s})
\end{equation}
\begin{equation}
F_{l}' = \mathrm{BatchNorm}(F_{l})
\end{equation}
where the Batch Normalization operation \(\mathrm{BatchNorm}(\cdot)\) standardizes feature maps across each batch, thus improving training stability.

Finally, these normalized feature maps are reshaped and transposed to form sequences of patch embeddings compatible as input to the Directional Attention Transformer. Mathematically, the reshaping and transposition operation is defined as:
\begin{equation}
F_{\text{scale}}'' \in \mathbb{R}^{B \times (H_{\text{scale}}W_{\text{scale}}) \times D},
\end{equation}
where \(\text{scale}\) denotes the specific scale (fine, small, or large), and \(H_{\text{scale}}, W_{\text{scale}}\) represent the corresponding spatial dimensions of each scale-specific embedding.

This multi-scale adaptive embedding strategy addresses the segmentation challenge of capturing both narrow, hairline cracks and wider crack formations within pavement images. By employing distinct yet complementary scale-specific convolutions, the SAE module ensures robust feature extraction at multiple resolutions, effectively supporting downstream modules for precise segmentation without manual annotations. In the next section, the Directional Attention Module will be discussed.

\subsection{Directional Attention Transformer}

\begin{figure}[H]
    \centering
    \includegraphics[width=1.0\textwidth]{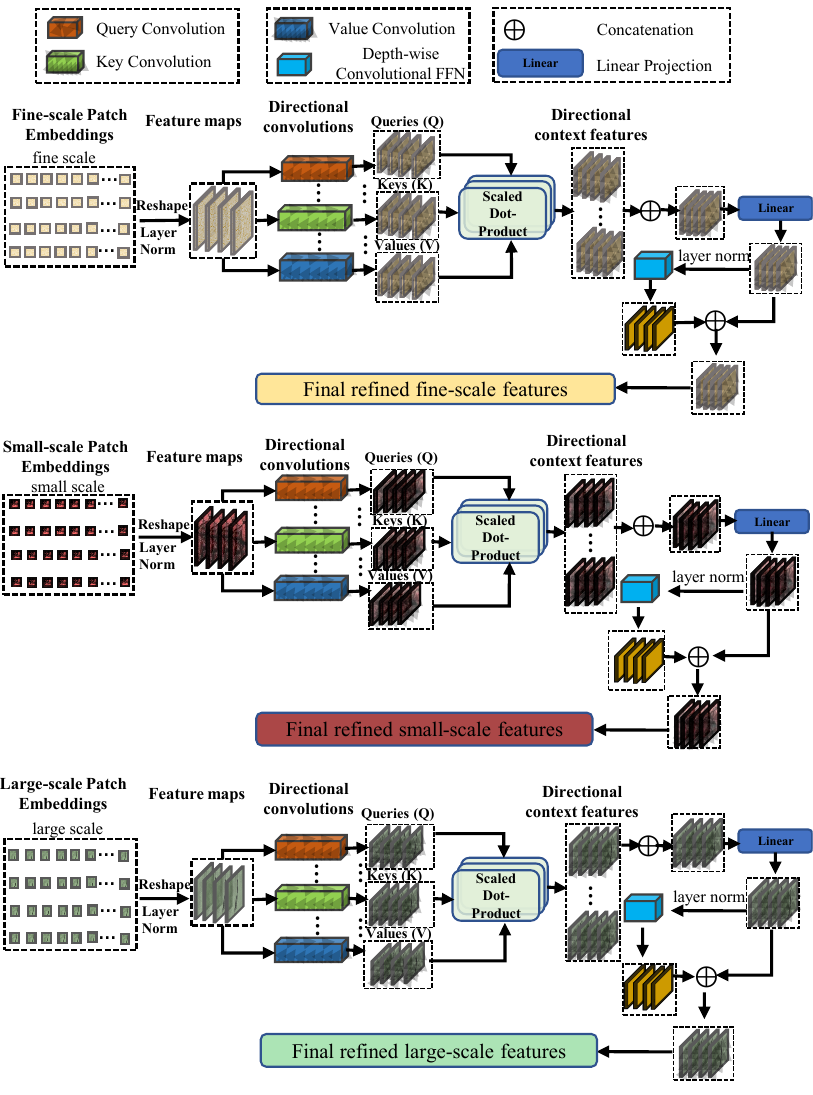}  % change file name and width as needed
    \caption{Directional Attention Transformer Module.}
    \label{fig:DAT module}
\end{figure}

The \textbf{Directional Attention Transformer (DAT)} module is designed to explicitly model directional spatial relationships, significantly enhancing the detection and segmentation of elongated, linear crack structures in pavement images. This module integrates multi‐scale embeddings from the Scale Adaptive Embedder (SAE). It refines feature representations using spatially‐directed attention mechanisms that effectively distinguish critical crack features from background noise. The architecture for the DAT module has been shown in Figure \ref{fig:DAT module}

The DAT module takes as input the multi‐scale embeddings from the SAE module, denoted as 
\[
F''_{\text{scale}} \in \mathbb{R}^{B \times (H_{\text{scale}}W_{\text{scale}}) \times D},
\]
where \(\text{scale}\in\{f,s,l\}\) corresponds to fine, small, and large scales respectively, \(B\) is the batch size, and \(D\) is the embedding dimension. These embeddings are first normalized via Layer Normalization to stabilize and improve training:

\begin{equation}
\hat{F}_{\text{scale}} 
= \mathrm{LayerNorm}\bigl(F''_{\text{scale}}\bigr),
\quad
\hat{F}_{\text{scale}}\in\mathbb{R}^{B\times(H_{\text{scale}}W_{\text{scale}})\times D}.
\end{equation}

Subsequently, the normalized embeddings are reshaped back to their spatial dimensions:

\begin{equation}
\hat{F}_{\text{scale}}^{\mathrm{spatial}}
\in\mathbb{R}^{B\times D\times H_{\text{scale}}\times W_{\text{scale}}}.
\end{equation}

Directional convolutions are then applied to capture elongated structural patterns essential for crack segmentation. These convolutions are defined as functions \(g\), parameterized by kernels \(W_k\) and biases \(b_k\). Specifically, for each direction \(k\) (e.g., horizontal \((1,3)\), vertical \((3,1)\)), the following is computed:

\begin{align}
Q_k &= g\bigl(\hat{F}_{\text{scale}}^{\mathrm{spatial}};W_{Q_k},b_{Q_k}\bigr),
\quad Q_k\in\mathbb{R}^{B\times D\times H_{\text{scale}}\times W_{\text{scale}}},\\
K_k &= g\bigl(\hat{F}_{\text{scale}}^{\mathrm{spatial}};W_{K_k},b_{K_k}\bigr),
\quad K_k\in\mathbb{R}^{B\times D\times H_{\text{scale}}\times W_{\text{scale}}},
\end{align}

where

\begin{equation}
g(X;W,b) = W \ast X + b,
\end{equation}

with \(\ast\) denoting convolution. Here \(W_{Q_k},W_{K_k}\in\mathbb{R}^{D\times D\times k_h\times k_w}\) are directional kernels and \(b_{Q_k},b_{K_k}\) are biases.

Values are obtained via point‐wise convolution:

\begin{equation}
V = W_V \ast \hat{F}_{\text{scale}}^{\mathrm{spatial}} + b_V,
\quad
V\in\mathbb{R}^{B\times D\times H_{\text{scale}}\times W_{\text{scale}}}.
\end{equation}

Attention maps are computed by softmax‐normalizing the element‐wise similarity of queries and keys:

\begin{equation}
A_k = \mathrm{softmax}\!\Bigl(\tfrac{Q_k \odot K_k}{\sqrt{D}}\Bigr),
\quad
A_k\in\mathbb{R}^{B\times D\times H_{\text{scale}}\times W_{\text{scale}}},
\end{equation}

where \(\odot\) is element‐wise multiplication. Directional context features follow:

\begin{equation}
C_k = A_k \odot V,
\quad
C_k\in\mathbb{R}^{B\times D\times H_{\text{scale}}\times W_{\text{scale}}}.
\end{equation}

All \(C_k\) are concatenated and reprojected:

\begin{equation}
F_{\text{scale}}^{\mathrm{attn}}
= W_O \bigl(C_1 \oplus C_2 \oplus \dots \oplus C_K\bigr) + b_O,
\quad
F_{\text{scale}}^{\mathrm{attn}}\in\mathbb{R}^{B\times D\times H_{\text{scale}}\times W_{\text{scale}}},
\end{equation}

where \(\oplus\) denotes channel‐wise concatenation.

Finally, these features undergo Layer Normalization and a depth‐wise convolutional feed‐forward network:

\begin{align}
\tilde{F}_{\text{scale}}
&= \mathrm{LayerNorm}\bigl(F_{\text{scale}}^{\mathrm{attn}}\bigr),\\
F_{\text{scale}}^{\mathrm{FFN}}
&= \mathrm{FFN}_{dw}\bigl(\tilde{F}_{\text{scale}}\bigr)
  + F_{\text{scale}}^{\mathrm{attn}},
\quad
F_{\text{scale}}^{\mathrm{FFN}}\in\mathbb{R}^{B\times D\times H_{\text{scale}}\times W_{\text{scale}}}.
\end{align}

The output \(F_{\text{scale}}^{\mathrm{FFN}}\) thus encodes directional spatial relationships, enhancing multi‐scale crack segmentation. These refined features feed into the Attention‐Guided Fusion module.

\subsection{Attention-Guided Fusion Module}

\begin{figure}[H]
    \centering
    \includegraphics[width=1.0\textwidth]{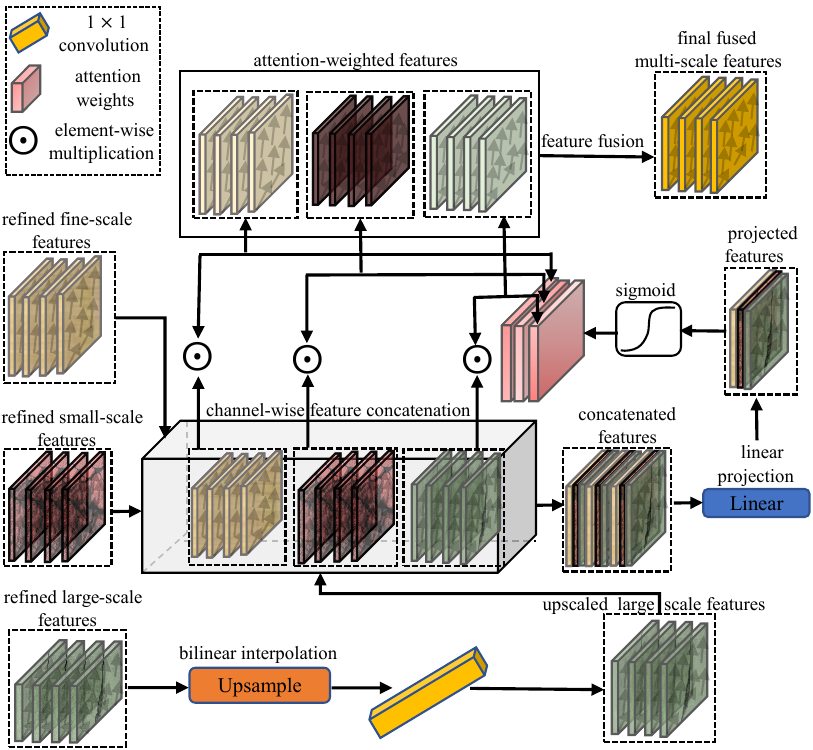}  % change file name and width as needed
    \caption{Attention-Guided Fusion Module.}
    \label{fig:AGF module}
\end{figure}

The \textbf{Attention-Guided Fusion (AGF)} module intelligently integrates multi-scale feature representations into a unified, robust feature map. Following the Scale Adaptive Embedder and Directional Attention Transformer modules, this module uses adaptive attention mechanisms to determine the optimal combination of features from the fine, small, and large scales. The AGF module applies dynamic weights to each scale-specific feature according to its contextual relevance. This ensures that the final representation captures essential pavement crack details across multiple resolutions. Figure \ref{fig:AGF module} shows the full architecture for the AGF module.

Let \(F_{f}^{\mathrm{FFN}} \in \mathbb{R}^{B \times D \times H \times W}\), \(F_{s}^{\mathrm{FFN}} \in \mathbb{R}^{B \times D \times H \times W}\), and \(F_{l}^{\mathrm{FFN}} \in \mathbb{R}^{B \times D \times \tfrac{H}{2} \times \tfrac{W}{2}}\) denote refined features from the fine, small, and large scales, respectively, obtained from the Directional Attention Transformer. To align spatial dimensions across scales, the large-scale feature map \(F_{l}^{\mathrm{FFN}}\) is first upsampled and projected:
\begin{equation}
F_{l}^{\mathrm{proj}}
=
\mathrm{Conv}_{1\times 1}\bigl(\mathrm{Upsample}(F_{l}^{\mathrm{FFN}})\bigr),
\quad
F_{l}^{\mathrm{proj}} \in \mathbb{R}^{B \times D \times H \times W},
\end{equation}
where \(\mathrm{Conv}_{1\times1}\) is a \(1\times1\) convolution to reduce dimensionality and match spatial dimensions, and \(\mathrm{Upsample}\) denotes bilinear interpolation.

Next, these scale-specific feature maps are concatenated along the channel dimension:
\begin{equation}
F_{\mathrm{cat}}
=
\bigl[F_{l}^{\mathrm{proj}};\,F_{s}^{\mathrm{FFN}};\,F_{f}^{\mathrm{FFN}}\bigr],
\quad
F_{\mathrm{cat}} \in \mathbb{R}^{B \times 3D \times H \times W}.
\end{equation}

The composite map \(F_{\mathrm{cat}}\) undergoes an attention-based weighting mechanism:
\begin{equation}
A
=
\sigma\bigl(W_{A}\ast F_{\mathrm{cat}} + b_{A}\bigr),
\quad
A \in \mathbb{R}^{B \times 3 \times H \times W},
\end{equation}
where \(W_{A}\in\mathbb{R}^{3\times 3D\times 1\times1}\), \(b_{A}\in\mathbb{R}^{3}\), and \(\sigma\) is the sigmoid activation. \(A\) provides scale-specific attention weights.

The concatenated feature map (\(F_{\mathrm{cat}}\)) is then split back into its scale-specific components (\(F_{l}^{\mathrm{split}}\), \(F_{s}^{\mathrm{split}}\), \(F_{f}^{\mathrm{split}}\)): 
\begin{align}
F_{l}^{\mathrm{split}} &\in \mathbb{R}^{B \times D \times H \times W}, \\
F_{s}^{\mathrm{split}} &\in \mathbb{R}^{B \times D \times H \times W}, \\
F_{f}^{\mathrm{split}} &\in \mathbb{R}^{B \times D \times H \times W}.
\end{align}

Each component is weighted by its corresponding attention slice:
\begin{align}
F_{l}^{\mathrm{weighted}}
&= F_{l}^{\mathrm{split}}\;\odot\;A[:,0:1,:,:], \\
F_{s}^{\mathrm{weighted}}
&= F_{s}^{\mathrm{split}}\;\odot\;A[:,1:2,:,:], \\
F_{f}^{\mathrm{weighted}}
&= F_{f}^{\mathrm{split}}\;\odot\;A[:,2:3,:,:],
\end{align}
where \(\odot\) denotes element-wise multiplication.

Finally, the attention-weighted features are summed:
\begin{equation}
F_{\mathrm{fused}}
=
F_{l}^{\mathrm{weighted}}
+
F_{s}^{\mathrm{weighted}}
+
F_{f}^{\mathrm{weighted}},
\quad
F_{\mathrm{fused}} \in \mathbb{R}^{B \times D \times H \times W}.
\end{equation}

This adaptive fusion strategy ensures multi-scale information is optimally integrated for accurate pavement crack segmentation.

\subsection{Cross-scale Consistency Loss}
To effectively address the challenge of self-supervised pavement crack segmentation, the framework incorporates a cross-scale consistency loss \cite{pmlr-v227-karimijafarbigloo24a}, composed of two complementary components: Inter-scale Consistency Loss and Intra-scale Consistency Loss. These losses are specifically designed to promote consistency and coherence in the learned feature representations across and within scales, significantly enhancing the robustness and accuracy of the model without reliance on manual annotations.

\subsubsection{Inter-scale Consistency Loss}

The Inter-scale Consistency Loss enforces similarity by minimizing the cosine distance between feature representations at different scales (fine, small, large) , encouraging the model to learn scale-invariant crack features. Let \(\mathbf{G}^{f},\mathbf{G}^{s},\mathbf{G}^{l}\in\mathbb{R}^{d}\) be the contextual feature vectors from fine, small, and large scales, respectively. The cosine similarity between two vectors is defined as:
\begin{equation}
\cos\bigl(\mathbf{G}^{x},\mathbf{G}^{y}\bigr)
=
\frac{\mathbf{G}^{x}\cdot \mathbf{G}^{y}}
     {\|\mathbf{G}^{x}\|\;\|\mathbf{G}^{y}\|},
\qquad
x,y\in\{f,s,l\}.
\end{equation}
The Inter-scale Consistency Loss \(\mathcal{L}_{\mathrm{inter}}(\mathbf{G}^{x},\mathbf{G}^{y})\) minimizes the cosine dissimilarity:
\begin{equation}
\mathcal{L}_{\mathrm{inter}}(\mathbf{G}^{x},\mathbf{G}^{y})
=
1 \;-\;\cos\bigl(\mathbf{G}^{x},\mathbf{G}^{y}\bigr).
\end{equation}
In practice, this loss is computed pairwise between fine–small and small–large scales, with a weighting factor:
\begin{equation}
\mathcal{L}_{\mathrm{inter}}
=
\lambda_{1}\Bigl[
  \mathcal{L}_{\mathrm{inter}}(\mathbf{G}^{f},\mathbf{G}^{s})
  +
  \mathcal{L}_{\mathrm{inter}}(\mathbf{G}^{s},\mathbf{G}^{l})
\Bigr],
\end{equation}
where \(\lambda_{1}\) is the weighting factor for the inter-scale consistency loss.

\subsubsection{Intra-scale Consistency Loss}
The Intra-scale Consistency Loss improves internal consistency within each scale-specific feature representation. Given an attention map \(\mathbf{A}\in\mathbb{R}^{L\times L}\) and the identity matrix \(\mathbf{I}\in\mathbb{R}^{L\times L}\), \(\mathbf{A}\) is enforced to resemble \(\mathbf{I}\) via an \(L_{1}\) loss:
\begin{equation}
\mathcal{L}_{\mathrm{intra}}(\mathbf{A})
=
\frac{1}{L^{2}}
\sum_{i=1}^{L}\sum_{j=1}^{L}
\bigl|A_{ij}-I_{ij}\bigr|.
\end{equation}
The loss is weighted as:
\begin{equation}
\mathcal{L}_{\mathrm{intra}}
=
\lambda_{2}\;\mathcal{L}_{\mathrm{intra}}(\mathbf{A}),
\end{equation}
where \(\lambda_{2}\) is the weighting factor for the intra‐scale consistency loss.

\subsubsection{Cross-Entropy Loss}
To further support self-supervision, a target was derived from the model's output itself \(\mathbf{T} = [T_{n}]\in\{0,1\}^{N}\) by taking the highest predicted crack probability for each pixel:
\begin{equation}
T_{n} =
\begin{cases}
1, & \text{if }O_{n} \ge 0.5,\\
0, & \text{otherwise},
\end{cases}
\quad n = 1,\dots,N,
\end{equation}
where \(\mathbf{O} = [O_{n}]\in[0,1]^{N}\) is the model’s output probability map and \(N\) is the total number of pixels.

The binary Cross-Entropy Loss between \(\mathbf{O}\) and \(\mathbf{T}\) is then defined as
\begin{equation}
\mathcal{L}_{\mathrm{CE}}(\mathbf{O},\mathbf{T})
=
-\frac{1}{N}
\sum_{n=1}^{N}
\Bigl[T_{n}\,\log\bigl(O_{n}\bigr)
+(1-T_{n})\,\log\bigl(1 - O_{n}\bigr)\Bigr],
\end{equation}
where \(\mathcal{L}_{\mathrm{CE}}\) denotes the cross-entropy loss, \(O_{n}\) is the predicted probability that pixel \(n\) belongs to the crack class, and \(T_{n}\) is the corresponding pseudo-target label.

\subsubsection{Total Consistency Loss}
Finally, the overall cross-scale consistency loss integrates all components:
\begin{equation}
\mathcal{L}_{\mathrm{total}}
=
\mathcal{L}_{\mathrm{CE}}(\mathbf{O},\mathbf{T})
+
\frac{1}{B}
\sum_{b=1}^{B}
\Bigl(
  \mathcal{L}_{\mathrm{inter}}^{(b)}
  +
  \mathcal{L}_{\mathrm{intra}}^{(b)}
\Bigr),
\end{equation}
where \(B\) is the batch size.

These self-supervisory losses ensure multi-scale representations remain coherent, internally consistent, and aligned with the model’s predictions, significantly enhancing segmentation accuracy without ground truth annotations (masks).

\section{Experimental details}
In this section, the ten public datasets and the preprocessing steps used to prepare data for training and testing are described. Each dataset includes unique crack patterns and imaging conditions, which enables testing the model across varied scenarios. The routine adopted for training the proposed model including the  training settings and evaluation metrics is then explained. The study is complemented with ablation studies that helps isolate the contribution of each module in the proposed model. 

\subsection{Datasets and Preprocessing}
\label{sec:datasets}
Experiments were conducted on ten publicly available datasets to assess the effectiveness of the proposed self supervised segmentation approach. Each dataset contains images capturing different crack patterns across various surfaces, lighting conditions, and environmental scenarios. Utilizing these diverse datasets allows effective assessment of the generalization capability and robustness of the segmentation model under different real-world conditions. Specifically, the following datasets were employed: CFD dataset \cite{CFD}, Crack500 \cite{CRACK500}, CrackTree200 \cite{cracktree200}, DeepCrack \cite{deepcrack}, Eugen Miller \cite{eugen_miller}, Forest \cite{forest}, GAPs384 \cite{GAPS384}, Rissbilder \cite{Volker}, Sylvie \cite{sylvie}, and Volker \cite{Volker}. Each dataset is characterized by distinct types of cracks, surface materials, and varying illumination conditions, making them well-suited for training and validating the self-supervised model's performance. Table~\ref{tab:crack_datasets_simple} provides a concise overview of each dataset's key characteristics, highlighting the variations in crack patterns and imaging environments. 

Furthermore, each dataset was split into training and testing subsets, maintaining an 80:20 split to ensure consistency in the evaluation protocol. This division provides a fair comparison of the model's predictive performance on previously unseen data. To enhance the generalization and robustness of the proposed model model, several data augmentation strategies were applied during training. These augmentations included random horizontal and vertical flips, rotation transformations, and scaling variations. These augmentation techniques simulate the variability encountered in practical pavement inspection scenarios, enabling the model to better generalize and accurately segment cracks under diverse conditions.

\begin{table}[H]
  \centering
  \begin{tabularx}{\textwidth}{l Y Y Y}
    \hline
    \textbf{Dataset}      & \textbf{Crack Types}                     & \textbf{Surface Material}              & \textbf{Lighting Conditions}             \\ \hline
    CFD          & Thin linear cracks              & Asphalt pavement              & Outdoor daylight, shadows, oil stains \\ 
    \hline
    Crack500     & Hairline, wide cracks                  & Asphalt road surfaces         & Mixed outdoor, varied weather         \\
    \hline
    CrackTree200 & Linear, alligator cracks        & Asphalt pavement              & Low contrast, uneven lighting         \\
    \hline
    DeepCrack    & Pavement, stone cracks          & Asphalt concrete; stone       & Daylight, some laser-lit               \\
    \hline
    Eugen Miller & Random cracks                   & Tunnel concrete               & Tunnel lighting                        \\
    \hline
    Forest       & Thin linear cracks              & Asphalt pavement              & Outdoor daylight, shadows              \\
    \hline
    GAPs384      & Longitudinal, transverse, block & Asphalt roads                 & Dry daylight                           \\
    \hline
    Rissbilder   & Architectural cracks            & Concrete, masonry             & Varied lighting                        \\
    \hline
    Sylvie       & Linear, network cracks          & Asphalt pavement              & Outdoor varied lighting                \\
    \hline
    Volker       & Structural cracks               & Concrete facades              & Field conditions, well-lit             \\ \hline
  \end{tabularx}
  \caption{Comparison of crack detection datasets}
  \label{tab:crack_datasets_simple}
\end{table}

\subsection{Implementation details}
The proposed self-supervised segmentation model was developed using the PyTorch deep learning library. The AdamW optimizer was used with a weight decay of \(1\times10^{-5}\) to mitigate potential overfitting. The initial learning rate was established at \(1\times10^{-4}\). To enhance training efficiency, a learning rate scheduler that reduced the learning rate by a factor of 0.5 whenever validation performance stagnated for five consecutive epochs was used.

Only the original images from the datasets were used during training, explicitly excluding any segmentation masks to ensure a genuinely self supervised learning scenario.. For all experiments, the datasets were randomly divided into training and validation subsets, ensuring consistency and fairness across evaluations. For model evaluation and performance benchmarking, predicted segmentation masks were directly compared against the ground truth masks available within the validation datasets.

Training was performed with a batch size of 8 for up to 500 epochs. To prevent overfitting and reduce unnecessary computation, an early stopping mechanism was implemented during training. Specifically, the total validation loss comprising the cross-entropy loss and the cross-scale consistency losses (inter-scale and intra-scale) was monitored at the end of each epoch. Training was automatically halted if no improvement in this total validation loss was observed for 100 consecutive epochs. The choice of 100 as the patience value was guided by empirical testing across different datasets, where it consistently offered a good trade-off between allowing sufficient optimization time and avoiding prolonged training after convergence. This early stopping configuration was uniformly applied across all experiments and baseline comparisons to maintain consistency.

The performance of the proposed model was compared with 14 state-of-the-art fully supervised segmentation models, including FCN \cite{FCN}, U-Net \cite{Unet}, U-Net++ \cite{Unet++}, PSPNet \cite{PSPNet}, PAN \cite{PAN}, MAnet \cite{MAnet}, LinkNet \cite{Linknet}, FPN \cite{FPN}, DeepLabV3 \cite{chen2017rethinking}, DeepLabV3+ \cite{deeplabv3plus}, UPerNet \cite{UPerNet}, Segformer \cite{Segformer}, and CrackFormer \cite{CrackFormer}. In total, all 14 segmentation models (the 13 fully-supervised baselines and the proposed self-supervised model) were each trained on all the 10 datasets, yielding a total of 140 experimental runs. Each baseline model was trained using the same experimental settings as the proposed model for consistency.

All computations were performed using the hardware and software configurations detailed in Table~\ref{tab:hardware_software}.

\begin{table}[h]
    \centering
    \begin{tabular}{ll}
        \hline
        \textbf{Component} & \textbf{Details} \\ \hline
        GPU                    & NVIDIA A40 (48\,GB) \\
        Framework              & PyTorch 2.7 \\
        Programming Language   & Python 3.9.12 \\
        CUDA Version           & 11.8 \\
        Optimizer              & AdamW \\
        Learning Rate Scheduler& Adaptive learning rate annealing \\ \hline
    \end{tabular}
    \caption{Summary of the hardware and software environment used in experiments.}
    \label{tab:hardware_software}
\end{table}

\subsection{Evaluation metrics}
Evaluation of the models were conducted using several standard metrics, including  mean Intersection over Unio (mIoU) and Dice coefficient. While these metrics assess overall overlap and similarity, additional metrics such as the XOR metric and Hausdorff Distance (HD) were incorporated to further capture spatial disagreement and misalignment, providing deeper insights into the segmentation quality. 

\paragraph{\textbf{Mean Intersection over Union}} The mIoU metric calculates the average overlap between predicted masks and ground truth masks across different classes. For a particular class, IoU is computed by dividing the intersection of the prediction and ground truth by their union:

\[
\text{IoU}_c = \frac{|P_c \cap G_c|}{|P_c \cup G_c|},
\]

where $P_c$ and $G_c$ represent the predicted and ground truth pixel sets for class $c$. The mIoU then averages the IoU scores across all $C$ classes:

\[
\text{mIoU} = \frac{1}{C} \sum_{c=1}^{C} \text{IoU}_c.
\]

\paragraph{\textbf{Dice Coefficient}} The Dice coefficient measures how closely the predicted segmentation aligns with the ground truth. It emphasizes regions with smaller or finer details, making it particularly useful for pavement cracks. Dice is computed as:

\[
\text{Dice} = \frac{2 \times |P \cap G|}{|P| + |G|},
\]

where $P$ and $G$ denote the predicted and actual sets of crack pixels, respectively.

\paragraph{\textbf{XOR}}  
The XOR metric quantifies discrepancies between the predicted and ground truth masks. It highlights areas exclusively classified as crack or non‐crack in one mask but not in the other, thereby capturing mismatches effectively:
\begin{equation*}
\mathrm{XOR} \;=\; \frac{1}{H\,W} \sum_{i=1}^{H} \sum_{j=1}^{W} \bigl(P_{ij} \oplus G_{ij}\bigr),
\end{equation*}
where \(P_{ij}\) and \(G_{ij}\) are the predicted and ground truth binary labels at pixel \((i,j)\), and \(\oplus\) denotes the logical XOR operation. A lower XOR value indicates better segmentation performance.

\paragraph{\textbf{Hausdorff Distance}}  
The Hausdorff Distance evaluates the spatial dissimilarity between predicted and ground truth masks. It quantifies the extent of spatial misalignment or inconsistency, defined mathematically as:
\begin{equation*}
\mathrm{HD}(P, G) \;=\; \max \Biggl\{\, 
\max_{p \in P} \min_{g \in G} \bigl\lVert p - g \bigr\rVert_{2},
\;\;
\max_{g \in G} \min_{p \in P} \bigl\lVert g - p \bigr\rVert_{2}
\Biggr\},
\end{equation*}
where \(P\) and \(G\) are the sets of crack‐pixel coordinates in the predicted and ground truth masks, respectively, and \(\lVert \cdot \rVert_{2}\) denotes the Euclidean distance. Lower HD values indicate higher segmentation accuracy, emphasizing better spatial alignment between prediction and ground truth.

\section{Results and Discussion}
The study reports the performance of the full Crack-Segmenter model along with ablated variants labeled v0, v1, v2, v3  in Table~\ref{tab:combined_results_all}. These variants represent simplified versions of the model with certain modules removed. Their architectural configurations are also described in Section~5.2. 
\subsection{Quantitative results}
Table~\ref{tab:combined_results_all} presents a comprehensive comparison between the proposed Crack-Segmenter and thirteen fully supervised baselines across ten public crack datasets. Evaluation is based on four metrics: mIoU, Dice score, XOR, and HD metric.

The proposed model achieved significantly superior performance compared to existing methods across nearly all datasets. On the CFD dataset, the model recorded a remarkable mIoU of 0.8875 and a Dice score of 0.9340, substantially surpassing U-Net++ (the best competitor among supervised methods) with an mIoU of 0.5257 and Dice of 0.6869. Moreover, the method exhibited notably lower XOR (0.6138) and HD (0.6138) values, demonstrating minimal spatial discrepancies and better alignment of predicted cracks.

Similarly, on the CRACK500 dataset, the self-supervised model achieved outstanding results, with an mIoU of 0.9332 and Dice of 0.9647, greatly exceeding the best-performing supervised method, Linknet, which attained mIoU and Dice scores of 0.6449 and 0.7838, respectively. The XOR and HD values of 0.1957 and 0.1629 further confirmed the model's exceptional spatial accuracy and lower false-positive rates.

The DeepCrack and Forest datasets also highlighted the model's robustness, achieving mIoU values of 0.8217 and 0.8167, respectively. These values substantially outperformed the next best methods, U-Net++ (0.7166 mIoU on DeepCrack) and U-Net (0.5392 mIoU on Forest). The high Dice scores of 0.8952 (DeepCrack) and 0.8896 (Forest) reinforced the accuracy improvements enabled by the framework's modules.

The CrackTree200 dataset, known for challenging crack patterns, demonstrated a significant performance leap with Crack-Segmenter attaining an mIoU of 0.8670 and a Dice score of 0.9223, far superior to the second-best, U-Net, with 0.4861 mIoU and 0.6498 Dice scores. Similarly, Crack-Segmenter excelled on the GAPs dataset, delivering an mIoU of 0.8096 and Dice of 0.8854, surpassing Segformer’s mIoU of 0.4016 and Dice of 0.5714 by a large margin.

\begin{table}
\centering
\caption{Validation Results of the proposed model and other models across all datasets.}
\label{tab:combined_results_all}

%--------------------------------------------------------------------
% Sub-table 1: CFD, CRACK500, DeepCrack, forest (4 datasets)
%--------------------------------------------------------------------
\resizebox{\textwidth}{!}{%
\begin{tabular}{l
    cccc % CFD
    cccc % CRACK500
    cccc % DeepCrack
    cccc % forest
}
\toprule
& \multicolumn{4}{c}{\textbf{CFD}} 
& \multicolumn{4}{c}{\textbf{CRACK500}}
& \multicolumn{4}{c}{\textbf{DeepCrack}}
& \multicolumn{4}{c}{\textbf{Forest}} \\
\cmidrule(lr){2-5} \cmidrule(lr){6-9} \cmidrule(lr){10-13} \cmidrule(lr){14-17}
\textbf{Model} 
& \textbf{mIoU}$\uparrow$ & \textbf{Dice}$\uparrow$ & \textbf{XOR}$\downarrow$ & \textbf{HD}$\downarrow$ 
& \textbf{mIoU}$\uparrow$ & \textbf{Dice}$\uparrow$ & \textbf{XOR}$\downarrow$ & \textbf{HD}$\downarrow$ 
& \textbf{mIoU}$\uparrow$ & \textbf{Dice}$\uparrow$ & \textbf{XOR}$\downarrow$ & \textbf{HD}$\downarrow$ 
& \textbf{mIoU}$\uparrow$ & \textbf{Dice}$\uparrow$ & \textbf{XOR}$\downarrow$ & \textbf{HD}$\downarrow$ \\
\midrule
\textbf{FCN} 
& 0.3509 & 0.5155 & 1.7027 & 0.6491 
& 0.5750 & 0.7088 & 0.8035 & 0.4250 
& 0.6123 & 0.7448 & 0.7623 & 0.3877
& 0.3796 & 0.5460 & 1.5054 & 0.6204 \\ %done
\textbf{Segformer} 
& 0.4734 & 0.6398 & 1.6703 & 0.6421 
& 0.6381 & 0.7776 & 0.6747 & 0.4249 
& 0.6556 & 0.7910 & 0.6538 & 0.3694 
& 0.4743 & 0.6406 & 1.4148 & 0.6110 \\
\textbf{PSPNet} 
& 0.3552 & 0.5191 & 1.2426 & 0.6695 
& 0.6207 & 0.7653 & 0.6716 & 0.4197 
& 0.6576 & 0.7925 & 1.0336 & 0.4658 
& 0.3840 & 0.5487 & 1.2346 & 0.6523 \\
\textbf{Linknet} 
& 0.4648 & 0.6232 & 1.7669 & 0.6835 
& 0.6449 & 0.7838 & 0.6987 & 0.4263 
& 0.7047 & 0.8243 & 1.3199 & 0.5122 
& 0.4883 & 0.6412 & 1.5864 & 0.6554 \\
\textbf{FPN} 
& 0.4212 & 0.5833 & 1.3548 & 0.6373 
& 0.6316 & 0.7740 & 0.6503 & 0.4348 
& 0.6783 & 0.8078 & 0.9757 & 0.4544 
& 0.4343 & 0.5976 & 1.2351 & 0.6181 \\
\textbf{Unet} 
& 0.5123 & 0.6743 & 2.2378 & 0.7018 
& 0.6398 & 0.7796 & 0.7171 & 0.4174 
& 0.7059 & 0.8255 & 1.2395 & 0.4970 
& 0.5392 & 0.6974 & 1.5924 & 0.6441 \\
\textbf{PAN} 
& 0.3654 & 0.5166 & 78.3467 & 0.9853 
& 0.6231 & 0.7674 & 0.6937 & 0.4240 
& 0.6535 & 0.7891 & 0.8343 & 0.4317 
& 0.4580 & 0.6261 & 1.2560 & 0.6160 \\
\textbf{DeepLabV3} 
& 0.3167 & 0.4744 & 1.9940 & 0.7353 
& 0.6420 & 0.7816 & 0.6699 & 0.4132 
& 0.6807 & 0.8092 & 1.0298 & 0.4622 
& 0.4821 & 0.6477 & 1.3812 & 0.6352 \\
\textbf{DeepLabV3Plus} 
& 0.3863 & 0.5526 & 1.3633 & 0.6674 
& 0.6376 & 0.7784 & 0.7092 & 0.4166 
& 0.6641 & 0.7961 & 0.9354 & 0.4437 
& 0.4610 & 0.6273 & 1.2988 & 0.6079 \\
\textbf{CrackFormer} 
& 0.4494 & 0.6178 & 1.4242 & 0.6186 
& 0.6306 & 0.7702 & 0.8168 & 0.4435 
& 0.6179 & 0.7592 & 0.6708 & 0.3755 
& 0.4450 & 0.6139 & 1.3694 & 0.6201 \\
\textbf{UPerNet} 
& 0.4674 & 0.6352 & 1.5864 & 0.6286 
& 0.6432 & 0.7813 & 0.7292 & 0.4255 
& 0.6581 & 0.7929 & 0.6514 & 0.3644 
& 0.4724 & 0.6381 & 1.4423 & 0.6179 \\
\textbf{MAnet} 
& 0.5166 & 0.6761 & 1.6948 & 0.6589 
& 0.6341 & 0.7754 & 0.7453 & 0.4284 
& 0.7004 & 0.8216 & 1.3568 & 0.5158 
& 0.5174 & 0.6752 & 1.6298 & 0.6495 \\
\textbf{UnetPlusPlus} 
& 0.5257 & 0.6869 & 1.7524 & 0.6635 
& 0.6443 & 0.7834 & 0.6703 & 0.4202 
& 0.7166 & 0.8344 & 1.2447 & 0.4983 
& 0.5457 & 0.7044 & 1.5658 & 0.6388 \\
\toprule
\textbf{Crack-Segmenter (ours)} 
& \textbf{\textcolor{red}{0.8875}} & \textbf{\textcolor{red}{0.9340}} & \textbf{\textcolor{red}{0.6138}} & \textbf{\textcolor{red}{0.6138}} 
& \textbf{\textcolor{red}{0.9332}} & \textbf{\textcolor{red}{0.9647}} & \textbf{\textcolor{red}{0.1957}} & \textbf{\textcolor{red}{0.1629}} 
& \textbf{\textcolor{red}{0.8217}} & \textbf{\textcolor{red}{0.8952}} & \textbf{\textcolor{red}{0.3685}} & \textbf{\textcolor{red}{0.2405}} 
& \textbf{\textcolor{red}{0.8167}} & \textbf{\textcolor{red}{0.8896}} & \textbf{\textcolor{red}{0.5836}} & \textbf{\textcolor{red}{0.4003}} \\
\textbf{Crack-Segmenter-v0 (ours)} 
& 0.5435 & 0.7025 & 0.8597 & 0.4565 
& 0.8572 & 0.9210 & 0.1782 & 0.1428
& 0.6955 & 0.7898 & 0.5003 & 0.3045
& 0.5723 & 0.7257 & 0.7695 & 0.4277 \\
\textbf{Crack-Segmenter-v1 (ours)} 
& 0.5432 & 0.7023 & 0.8603 & 0.4568
& 0.5360 & 0.6965 & 0.8708 & 0.4640
& 0.5698 & 0.7026 & 0.7064 & 0.4303
& 0.5554 & 0.7102 & 0.7972 & 0.4446 \\
\textbf{Crack-Segmenter-v2 (ours)} 
& 0.5360 & 0.6965 & 0.8708  & 0.4640
& 0.5711 & 0.7247 & 0.7699 & 0.4289
& 0.6836 & 0.8013 & 0.5388 & 0.3164
& 0.5711 & 0.7247 & 0.7699 & 0.4289 \\
\textbf{Crack-Segmenter-v3 (ours)} 
& 0.5779 & 0.7302 & 0.7300 & 0.4221
& 0.7457 & 0.8407 & 0.3033 & 0.2543
& 0.6797 & 0.7996 & 0.5590 & 0.3203
& 0.5153 & 0.6617 & 0.7999 & 0.4847 \\
\bottomrule
\end{tabular}
}% end resizebox

\vspace{1em}

%--------------------------------------------------------------------
% Sub-table 2: cracktree200, GAPS384, Eugen\_Miller (3 datasets)
%--------------------------------------------------------------------
\resizebox{\textwidth}{!}{%
\begin{tabular}{l
    cccc % cracktree200
    cccc % GAPS384
    cccc % Eugen_Miller
}
\toprule
& \multicolumn{4}{c}{\textbf{CrackTree200}} 
& \multicolumn{4}{c}{\textbf{GAPs}} 
& \multicolumn{4}{c}{\textbf{Eugen Miller}} \\
\cmidrule(lr){2-5} \cmidrule(lr){6-9} \cmidrule(lr){10-13}
\textbf{Model} 
& \textbf{mIoU}$\uparrow$ & \textbf{Dice}$\uparrow$ & \textbf{XOR}$\downarrow$ & \textbf{HD}$\downarrow$ 
& \textbf{mIoU}$\uparrow$ & \textbf{Dice}$\uparrow$ & \textbf{XOR}$\downarrow$ & \textbf{HD}$\downarrow$ 
& \textbf{mIoU}$\uparrow$ & \textbf{Dice}$\uparrow$ & \textbf{XOR}$\downarrow$ & \textbf{HD}$\downarrow$ \\
\midrule 
\textbf{FCN} 
& 0.065 & 0.1215 & 13.8985 & 0.9353 
& 0.3648 & 0.5128 & 1.4705 & 0.6352 
& 0.6215 & 0.7660 & 0.5168 & 0.3785 \\
\textbf{Segformer} 
& 0.2916 & 0.4509 & 12.3826 & 0.9299 
& 0.4016 & 0.5714 & 1.3561 & 0.6268 
& 0.5973 & 0.7402 & 0.6956 & 0.4543 \\
\textbf{PSPNet} 
& 0.2930 & 0.4477 & 6.7798 & 0.9178 
& 0.2907 & 0.4432 & 1.2396 & 0.7015 
& 0.6062 & 0.7479 & 0.5292 & 0.3905 \\
\textbf{Linknet} 
& 0.4804 & 0.6411 & 12.2496 & 0.9355 
& 0.3996 & 0.5659 & 1.7396 & 0.6894 
& 0.6021 & 0.7414 & 0.5740 & 0.4178 \\
\textbf{FPN} 
& 0.2962 & 0.4469 & 7.6561 & 0.9197 
& 0.3068 & 0.4624 & 1.3438 & 0.7369 
& 0.4682 & 0.6161 & 0.6412 & 0.4997 \\
\textbf{Unet} 
& 0.4861 & 0.6498 & 11.4862 & 0.9301 
& 0.3832 & 0.5508 & 1.5567 & 0.6753 
& 0.6315 & 0.7655 & 0.4998 & 0.3771 \\
\textbf{PAN} 
& 0.2755 & 0.4284 & 6.7856 & 0.9185 
& 0.2761 & 0.4259 & 1.2340 & 0.7191 
& 0.5796 & 0.7265 & 0.5651 & 0.4189 \\
\textbf{DeepLabV3} 
& 0.3178 & 0.4708 & 8.4792 & 0.9221 
& 0.3685 & 0.5343 & 1.5615 & 0.6844 
& 0.6291 & 0.7633 & 0.5928 & 0.4039 \\
\textbf{DeepLabV3Plus} 
& 0.2788 & 0.4243 & 9.0466 & 0.9205 
& 0.3168 & 0.4705 & 1.5988 & 0.7019 
& 0.6047 & 0.7457 & 0.5501 & 0.3999 \\
\textbf{CrackFormer} 
& 0.2656 & 0.4189 & 12.3523 & 0.9348 
& 0.3470 & 0.5112 & 1.6896 & 0.6595 
& 0.4756 & 0.6352 & 0.8786 & 0.6001 \\
\textbf{UPerNet} 
& 0.2569 & 0.4049 & 12.9757 & 0.9328 
& 0.4029 & 0.5724 & 1.3435 & 0.6121 
& 0.5835 & 0.7295 & 0.6135 & 0.4350 \\
\textbf{MAnet} 
& 0.4198 & 0.5773 & 11.9556 & 0.9330 
& 0.3793 & 0.5414 & 1.6618 & 0.6844 
& 0.5892 & 0.7322 & 0.5547 & 0.4110 \\
\textbf{UnetPlusPlus} 
& 0.4255 & 0.5927 & 8.9129 & 0.9183 
& 0.3926 & 0.5561 & 1.6479 & 0.6858 
& 0.7072 & 0.8214 & 0.4798 & 0.3505 \\
\toprule
\textbf{Crack-Segmenter (ours)} 
& \textbf{\textcolor{red}{0.8670}} & \textbf{\textcolor{red}{0.9223}} & \textbf{\textcolor{red}{5.0154}} & \textbf{\textcolor{red}{0.8409}} 
& \textbf{\textcolor{red}{0.8096}} & \textbf{\textcolor{red}{0.8854}} & \textbf{\textcolor{red}{0.6542}} & \textbf{\textcolor{red}{0.4497}} 
& 0.8451 & 0.9071 & 0.3952 & 0.3669 \\
\textbf{Crack-Segmenter-v0 (ours)} 
& 0.1572 & 0.2716 & 5.1871 & 0.8428
& 0.6353 & 0.7711 & 0.6264 & 0.3647
& \textbf{\textcolor{red}{0.8597}} & \textbf{\textcolor{red}{0.9245}} & \textbf{\textcolor{red}{0.1635}} & \textbf{\textcolor{red}{0.1403}} \\
\textbf{Crack-Segmenter-v1 (ours)} 
& 0.1474 & 0.2568 & 5.8197 & 0.8526
& 0.6312 & 0.7682 & 0.6293 & 0.3688
& 0.8016 & 0.8860 & 0.2237 & 0.1984 \\
\textbf{Crack-Segmenter-v2 (ours)} 
& 0.1406 & 0.2453 & 5.1946 & 0.8594
& 0.6353 & 0.7711 & 0.6264 & 0.3647
& 0.6860 & 0.8080 & 0.3556 & 0.3140 \\
\textbf{Crack-Segmenter-v3 (ours)} 
& 0.1385 & 0.2430 & 5.0391 & 0.8615
& 0.4953 & 0.6450 & 0.7485 & 0.5047
& 0.7230 & 0.8369 & 0.3078 & 0.2770 \\
\bottomrule
\end{tabular}
}% end resizebox

\vspace{1em}

%--------------------------------------------------------------------
% Sub-table 3: Rissbilder, Sylvie, Volker (3 datasets)
%--------------------------------------------------------------------
\resizebox{\textwidth}{!}{%
\begin{tabular}{l
    cccc % Rissbilder
    cccc % Sylvie
    cccc % Volker
}
\toprule
& \multicolumn{4}{c}{\textbf{Rissbilder}} 
& \multicolumn{4}{c}{\textbf{Sylvie}} 
& \multicolumn{4}{c}{\textbf{Volker}} \\
\cmidrule(lr){2-5} \cmidrule(lr){6-9} \cmidrule(lr){10-13}
\textbf{Model} 
& \textbf{mIoU}$\uparrow$ & \textbf{Dice}$\uparrow$ & \textbf{XOR}$\downarrow$ & \textbf{HD}$\downarrow$ 
& \textbf{mIoU}$\uparrow$ & \textbf{Dice}$\uparrow$ & \textbf{XOR}$\downarrow$ & \textbf{HD}$\downarrow$ 
& \textbf{mIoU}$\uparrow$ & \textbf{Dice}$\uparrow$ & \textbf{XOR}$\downarrow$ & \textbf{HD}$\downarrow$ \\
\midrule
\textbf{FCN} 
& 0.5098 & 0.6707 & 0.8074 & 0.4902 
& 0.7498 & 0.8439 & 0.3016 & 0.2502 
& 0.6658 & 0.7981 & 0.4685 & 0.3342 \\
\textbf{Segformer} 
& 0.5169 & 0.6810 & 0.7938 & 0.4912 
& 0.6600 & 0.7899 & 0.3960 & 0.2939 
& 0.6632 & 0.7973 & 0.4732 & 0.3407 \\
\textbf{PSPNet} 
& 0.5042 & 0.6685 & 0.7935 & 0.5339 
& 0.5548 & 0.6998 & 0.4102 & 0.3592 
& 0.6865 & 0.8132 & 0.4625 & 0.3401 \\
\textbf{Linknet} 
& 0.6084 & 0.7497 & 0.8332 & 0.4941 
& 0.6430 & 0.7750 & 0.5250 & 0.3740 
& 0.7235 & 0.8344 & 0.4871 & 0.3439 \\
\textbf{FPN} 
& 0.5822 & 0.7348 & 0.7652 & 0.4932 
& 0.5993 & 0.7456 & 0.4893 & 0.3898 
& 0.7026 & 0.8242 & 0.4505 & 0.3341 \\
\textbf{Unet} 
& 0.6449 & 0.7836 & 0.8240 & 0.4836 
& 0.6558 & 0.7891 & 0.4501 & 0.3269 
& 0.7454 & 0.8532 & 0.4788 & 0.3349 \\
\textbf{PAN} 
& 0.5415 & 0.7015 & 0.7529 & 0.5039 
& 0.6073 & 0.7504 & 0.4030 & 0.3438 
& 0.6790 & 0.8076 & 0.4574 & 0.3405 \\
\textbf{DeepLabV3} 
& 0.5970 & 0.7474 & 0.7754 & 0.4857 
& 0.6639 & 0.7968 & 0.3633 & 0.2972
& 0.7213 & 0.8378 & 0.4448 & 0.3280 \\
\textbf{DeepLabV3Plus} 
& 0.5650 & 0.7210 & 0.7781 & 0.4965 
& 0.6508 & 0.7826 & 0.3683 & 0.3038 
& 0.6871 & 0.8139 & 0.4748 & 0.3467 \\
\textbf{CrackFormer} 
& 0.4638 & 0.6313 & 0.8590 & 0.5126 
& 0.4829 & 0.6402 & 0.6936 & 0.5363 
& 0.6182 & 0.7631 & 0.5042 & 0.3614 \\
\textbf{UPerNet} 
& 0.5150 & 0.6794 & 0.8045 & 0.4872 
& 0.6663 & 0.7946 & 0.3450 & 0.2680 
& 0.6682 & 0.8009 & 0.4628 & 0.3331 \\
\textbf{MAnet} 
& 0.6370 & 0.7771 & 0.8190 & 0.4811 
& 0.6254 & 0.7642 & 0.4140 & 0.3246 
& 0.7027 & 0.8245 & 0.5236 & 0.3598 \\
\textbf{UnetPlusPlus} 
& 0.6564 & 0.7920 & 0.7964 & 0.4744 
& 0.6719 & 0.8021 & 0.5109 & 0.3890 
& 0.7641 & 0.8659 & 0.4655 & 0.3274 \\
\toprule
\textbf{Crack-Segmenter (ours)} 
& 0.7998 & 0.8785 & 0.5047 & 0.4344
& 0.8891 & 0.9352 & 0.3341 & 0.3122
& 0.6707 & 0.7955 & 0.6434 & 0.6240 \\
\textbf{Crack-Segmenter-v0 (ours)} 
& 0.8075 & 0.8932 & 0.2398 & 0.1925 
& 0.9063 & 0.9503 & 0.1060 & 0.0937
& 0.8746 & 0.9330 & 0.1437 & 0.1254 \\
\textbf{Crack-Segmenter-v1 (ours)} 
& \textbf{\textcolor{red}{0.8079}} & \textbf{\textcolor{red}{0.8934}} & \textbf{\textcolor{red}{0.2394}} & \textbf{\textcolor{red}{0.1921}} 
&  \textbf{\textcolor{red}{0.9065}} & \textbf{\textcolor{red}{0.9504}} & \textbf{\textcolor{red}{0.1057}} & \textbf{\textcolor{red}{0.0935}}
& \textbf{\textcolor{red}{0.8761}} & \textbf{\textcolor{red}{0.9338}} & \textbf{\textcolor{red}{0.1421}} & \textbf{\textcolor{red}{0.1239}} \\
\textbf{Crack-Segmenter-v2 (ours)} 
& 0.7894 & 0.8810 & 0.2611 & 0.2106 
& 0.7873 & 0.8745 & 0.2326 & 0.2127
& 0.8637 & 0.9265 & 0.1560 & 0.1363\\

\textbf{Crack-Segmenter-v3 (ours)} 
& 0.6740 & 0.7979 & 0.3898 & 0.3260 
& 0.8802 & 0.9354 & 0.1344 & 0.1198
& 0.5804 & 0.7303 & 0.4637 & 0.4196 \\
\bottomrule
\end{tabular}
}% end resizebox

\end{table}

Performance on Eugen Miller was also notably superior, with Crack-Segmenter achieving an mIoU of 0.8451 and Dice of 0.9071, outperforming U-Net++ with 0.7072 mIoU and 0.8214 Dice. These results indicate the proposed model's effectiveness even in diverse structural conditions.

\begin{figure}[H]
    \centering
    \includegraphics[width=1.0\textwidth]{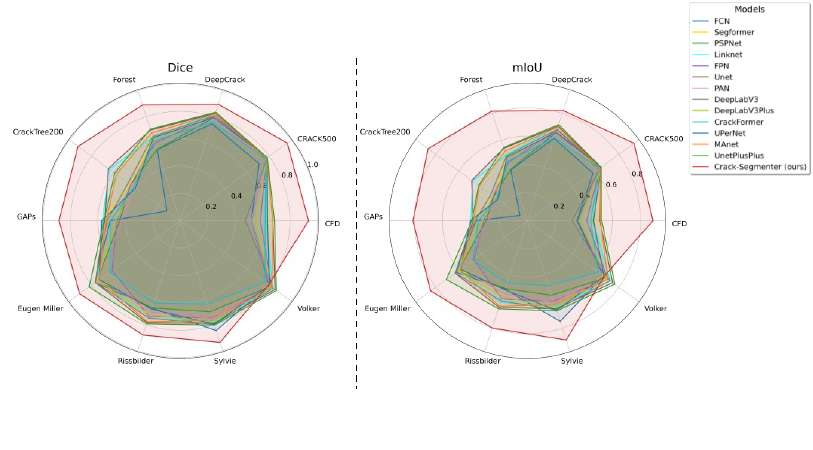}  % change file name and width as needed
    \caption{Radar plots for validation Dice and mIoU scores of all the baseline models and Crack-Segmenter across all the 10 datasets.}
    \label{fig:Radar plots}
\end{figure}

However, the performance on the Volker dataset was slightly lower, with an mIoU of 0.6707 and Dice of 0.7955, compared to U-NetPlusPlus, which achieved an mIoU of 0.7641 and Dice of 0.8659. This suggests room for further refinement of the multi-scale attention mechanisms to handle highly variable crack structures effectively. Figure \ref{fig:Radar plots} shows mIoU and Dice scores of Crack-Segmenter and all the baseline models across the different datasets summarised in a radar plot. 

The consistently superior quantitative results clearly demonstrate the effectiveness of the integrated modules within the proposed self-supervised segmentation framework. Specifically, the SAE module efficiently captures comprehensive multi-scale feature details; the Directional Attention Transformer emphasizes linear crack structures, improving the detection accuracy; and the Attention-Guided Fusion optimally merges multi-scale features, enhancing overall segmentation performance. Collectively, these modules facilitate accurate, annotation-free crack segmentation, significantly advancing the state-of-the-art in pavement distress assessment.

\subsection{Ablation studies}
To evaluate the individual contributions of each proposed module within the proposed self-supervised segmentation framework, the study conducted comprehensive ablation experiments on the DeepCrack dataset. This systematic analysis isolates the effectiveness of each component: the SAE module, DAT module, and AGF module, and examines their interactions when combined in various configurations.

\subsubsection{Experimental Setup}
The study evaluated six distinct architectural variants using mIoU and Dice score as primary metrics. Each variant represents a specific combination of the proposed modules, allowing quantification of individual contributions to segmentation performance.
The following variants were systematically evaluated:

\textbf{Baseline}: Standard U-Net architecture with ResNet-18 encoder, representing conventional supervised segmentation without self-supervised components.

\textbf{Crack-Segmenter}: Complete architecture incorporating all three proposed modules.

\textbf{Crack-Segmenter-v0}: Integration of SAE module only, focusing on multi-scale representation learning capabilities.

\textbf{Crack-Segmenter-v1}: Combination of SAE and DAT modules, emphasizing directional attention alongside multi-scale features.

\textbf{Crack-Segmenter-v2}: Integration of SAE and AGF modules, combining multi-scale embeddings with attention-guided fusion.

\textbf{Crack-Segmenter-v3}: Combination of DAT and AGF modules without multi-scale embeddings.

\subsubsection{Analysis}
To assess the individual contributions of each proposed modules, the study conducted systematic ablation experiments on the DeepCrack dataset. These studies are crucial for understanding the role each component plays in the overall performance of Crack-Segmenter. All ablation variants were evaluated using the four key metrics: mIoU, Dice score, XOR Score, and HD Score.

The analysis begins by defining the \textbf{Baseline} model, which consists of a standard U-Net architecture with a ResNet-18 encoder and none of the proposed self-supervised modules. This configuration serves as a conventional supervised benchmark. As shown in Table~\ref{tab:ablation}, it achieved an mIoU of 0.5866 and a Dice score of 0.7258. These results establish a foundational point of comparison for the proposed architectural variations.

 When only the SAE module was included in the architecture (\textbf{Crack-Segmenter-v0}), this configuration observed a substantial improvement in both metrics, reaching an mIoU of 0.6955 and a Dice score of  0.7898. This confirms the strong impact of multi-scale representations in capturing crack structures of varying widths. SAE provides the model with diverse spatial resolutions, enabling it to better recognize both fine-grained and coarse crack patterns, which are typically missed in single-scale encoders.

In the \textbf{Crack-Segmenter-v1} variant, DAT module was added on top of SAE while leaving out AGF. Interestingly, this combination led to a drop in performance (mIoU of 0.5698 and Dice of 0.7026), falling even below the baseline. This suggests that applying directional attention without a proper fusion mechanism may not be sufficient for effective feature integration. DAT focuses on enhancing linear crack continuity through spatially aware convolutions, but without adaptive fusion to resolve scale-level redundancies, it may introduce conflicting or misaligned representations.

The combination of SAE and AGF in \textbf{Crack-Segmenter-v2} demonstrated stronger performance, with 0.6836 mIoU and 0.8013 Dice. These results reinforce the importance of adaptive feature fusion when dealing with multi-scale representations. AGF dynamically learns to assign relevance weights across different spatial resolutions, ensuring optimal use of both local and contextual information during segmentation. The effective integration of scale-specific details allows the model to better adapt to irregular crack geometries and surrounding textures.

The study then evaluated \textbf{Crack-Segmenter-v3}, which includes only DAT and AGF, excluding SAE. This variant resulted in an mIoU of 0.6797 and Dice score of 0.7996. Although this is an improvement over the baseline, the absence of SAE resulted in weaker feature diversity, reducing the ability of the fusion and attention mechanisms to operate on rich, scale-aware representations. This again highlights the importance of SAE as a foundation for multi-scale learning.

Finally, the full model (\textbf{Crack-Segmenter}) that integrates all three modules: SAE, DAT, and AGF achieved the highest performance across all metrics: 0.8217 mIoU, 0.8952 Dice, 0.3685 XOR, and 0.2405 HD. These results confirm that the complete architecture benefits from the combined strengths of its components. SAE enables comprehensive multi-scale feature extraction, DAT enhances directional continuity, and AGF fuses these features adaptively. Together, this integration result in a robust and coherent segmentation map that generalizes well across complex crack structures and challenging visual conditions.

This observation becomes particularly important when analyzing model performance on the CrackTree200 dataset. It was observed that the CrackTree200 dataset exhibited the most pronounced performance gap between the full Crack-Segmenter and its ablated variants (v0–v3). Upon investigation, the study found that this dataset’s challenging visual properties such as low contrast, complex topology, and fine linear cracks demanded the combined contributions of SAE, DAT, and AGF. The ablated models, lacking one or more of these modules, were unable to capture these characteristics effectively, leading to steep performance drops. This result confirms the necessity of all three modules for effective generalization on particularly challenging datasets.

\begin{table}[H]
  \centering
  \caption{Ablation study results on the DeepCrack dataset. Each variant systematically evaluates different module combinations to assess individual contributions.}
  \label{tab:ablation}
  % reduce horizontal padding
  \setlength{\tabcolsep}{4pt}
  \small
  % scale down if wider than \linewidth
  \begin{adjustbox}{max width=\linewidth}
    \begin{tabular}{l|cccc|ccc}
      \toprule
      \multirow{2}{*}{\textbf{Model Variant}} 
        & \multicolumn{4}{c|}{\textbf{Metric}} 
        & \multicolumn{3}{c}{\textbf{Module Components}} \\
      \cmidrule(lr){2-5} \cmidrule(lr){6-8}
        & \textbf{mIoU}$\uparrow$ 
        & \textbf{Dice}$\uparrow$ 
        & \textbf{XOR}$\downarrow$ 
        & \textbf{HD}$\downarrow$ 
        & \textbf{SAE} 
        & \textbf{DAT} 
        & \textbf{AGF} \\
      \midrule
      Baseline               & 0.5866 & 0.7258 & 0.7675 & 0.4134 & \ding{55} & \ding{55} & \ding{55} \\
      Crack-Segmenter        & \textbf{0.8217} & \textbf{0.8952} & \textbf{0.3685} & \textbf{0.2405} & \ding{51} & \ding{51} & \ding{51} \\
      Crack-Segmenter-v0     & 0.6955 & 0.7898 & 0.5003 & 0.3405 & \ding{51} & \ding{55} & \ding{55} \\
      Crack-Segmenter-v1     & 0.5698 & 0.7026 & 0.7064 & 0.4303 & \ding{51} & \ding{51} & \ding{55} \\
      Crack-Segmenter-v2     & 0.6836 & 0.8013 & 0.5388 & 0.3164 & \ding{51} & \ding{55} & \ding{51} \\
      Crack-Segmenter-v3     & 0.6797 & 0.7996 & 0.5590 & 0.3203 & \ding{55} & \ding{51} & \ding{51} \\
      \bottomrule
    \end{tabular}
  \end{adjustbox}
\end{table}

These ablation results validate the design of the proposed self-supervised architecture and demonstrate how each module contributes to enhancing segmentation quality. The clear performance gains of the full model emphasize the necessity of combining multi-scale embeddings, directional attention, and guided fusion in an integrated manner for accurate pavement crack segmentation.

\subsection{Model Explainability}
Interpreting model behavior is essential for validating its reliability, especially in safety-critical applications such as pavement crack detection. To assess whether the proposed \textit{Crack-Segmenter} attends to relevant spatial structures, attention map visualizations were employed to analyze its focus during inference.

Attention maps were extracted from the final transformer block at each of the three spatial scales (fine, small, and large) in the architecture. These layers capture the model’s most refined representations across different resolutions. For blocks with multiple attention heads, attention weights were averaged across heads to produce a single map per scale, ensuring clarity while preserving dominant spatial patterns. All attention maps were then normalized to the range [0, 1] and overlaid as heatmaps on the original DeepCrack images, where color intensity reflects attention strength. Regions receiving strong attention appear warmer (e.g., red or yellow), while areas with low attention are cooler (e.g., blue), enabling clear visual identification of the model’s focus during segmentation.

\begin{figure}[h]
    \centering
    \includegraphics[width=1.0\textwidth]{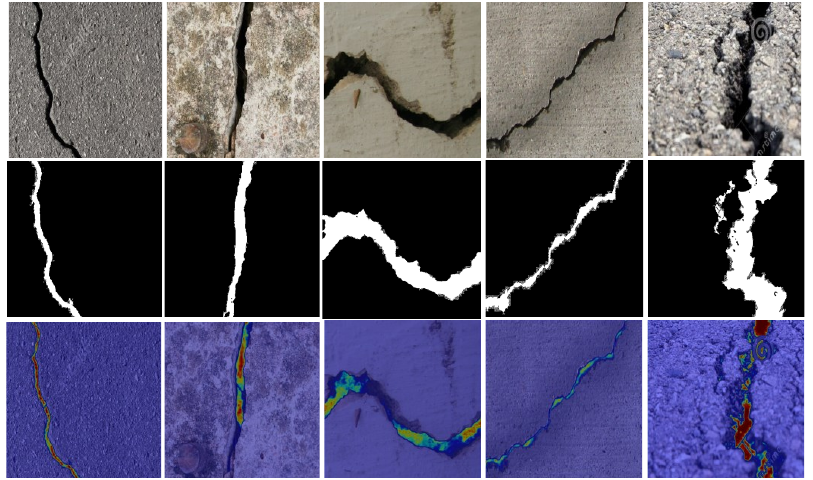}  % change file name and width as needed
    \caption{Attention map visualization of \textit{Crack-Segmenter} on samples images from DeepCrack dataset. Top row: input images. Middle row: predicted masks. Bottom row: attention map overlays highlighting regions the model focused on during crack segmentation.}
    \label{fig:attention_maps}
\end{figure}

Figure~\ref{fig:attention_maps} presents example attention visualizations for sample images in the DeepCrack dataset. The model consistently concentrates attention along actual crack regions, while suppressing distractors such as shadows, surface texture, and background noise. This focused behavior highlights the effectiveness of the Directional Attention Transformer and Attention-Guided Fusion modules in guiding the network toward semantically meaningful features. These visualizations confirm that the model captures relevant structural cues without explicit supervision, reinforcing both the design rationale and the effectiveness of the proposed framework.

\subsection{Statistical Analysis}
To rigorously assess the statistical significance of \textit{Crack-Segmenter}, the study conducted detailed statistical analysis comparing its performance against state-of-the-art segmentation methods. These analysis focused on evaluating consistency and superiority across multiple segmentation metrics, including  mIoU, Dice score, XOR, and HD. Table \ref{tab:model_performance} summarizes the mean performance and standard deviations of each method averaged over all datasets.

From Table \ref{tab:model_performance}, it is evident that the proposed \textit{Crack-Segmenter} achieved consistently superior results across all evaluation metrics. Specifically, the proposed model attained the highest mIoU (0.8340 ± 0.0714) and Dice score (0.9008 ± 0.0458), indicating significantly more accurate segmentation performance compared to other baseline models. Additionally, it showed the lowest XOR (0.9309 ± 1.4432) and HD (0.4446 ± 0.2013) values, reflecting fewer misclassifications and better spatial alignment with ground-truth crack structures.

\begin{table}[H]
\centering
\caption{Mean Performance and Standard Deviation by Model accross all datasets}
\label{tab:model_performance}
\resizebox{\textwidth}{!}{%
  \begin{tabular}{l|c|c|c|c}
    \hline
    \textbf{Model} & \textbf{mIoU}$^{\dagger}$ & \textbf{Dice}$^{\dagger}$ & \textbf{XOR}$^{\ddagger}$ & \textbf{HD}$^{\ddagger}$ \\
    \hline\hline
    DeepLabV3       & 0.5419 ± 0.1569 & 0.6863 ± 0.1434 & 1.7292 ± 2.4292 & 0.5367 ± 0.2004 \\
    DeepLabV3Plus   & 0.5252 ± 0.1525 & 0.6712 ± 0.1434 & 1.7123 ± 2.6091 & 0.5305 ± 0.1915 \\
    FCN             & 0.4894 ± 0.2017 & 0.6228 ± 0.2123 & 2.2237 ± 4.1299 & 0.5106 ± 0.2018 \\
    FPN             & 0.5121 ± 0.1475 & 0.6593 ± 0.1378 & 1.5562 ± 2.1700 & 0.5518 ± 0.1777 \\
    Linknet         & 0.5760 ± 0.1104 & 0.7180 ± 0.0930 & 2.1780 ± 3.5753 & 0.5532 ± 0.1849 \\
    MAnet           & 0.5722 ± 0.1111 & 0.7165 ± 0.0976 & 2.1355 ± 3.4871 & 0.5447 ± 0.1866 \\
    PAN             & 0.5059 ± 0.1529 & 0.6540 ± 0.1469 & 9.1329 ± 24.3942 & 0.5702 ± 0.2333 \\
    PSPNet          & 0.4953 ± 0.1525 & 0.6446 ± 0.1428 & 1.4397 ± 1.9048 & 0.5450 ± 0.1865 \\
    Segformer       & 0.5372 ± 0.1273 & 0.6880 ± 0.1146 & 2.0511 ± 3.6555 & 0.5184 ± 0.1894 \\
    UPerNet         & 0.5334 ± 0.1366 & 0.6829 ± 0.1266 & 2.0954 ± 3.8467 & 0.5105 ± 0.1941 \\
    Unet            & 0.5944 ± 0.1111 & 0.7369 ± 0.0923 & 2.1082 ± 3.3477 & 0.5388 ± 0.1948 \\
    UnetPlusPlus    & 0.6050 ± 0.1265 & 0.7439 ± 0.1050 & 1.8047 ± 2.5484 & 0.5366 ± 0.1867 \\
    CrackFormer     & 0.4796 ± 0.1187 & 0.6361 ± 0.1120 & 2.1258 ± 3.6134 & 0.5662 ± 0.1662 \\
    \hline
     \textbf{Crack‐Segmenter (Ours)} & \textbf{0.8340 ± 0.0714} & \textbf{0.9008 ± 0.0458} & \textbf{0.9309 ± 1.4432} & \textbf{0.4446 ± 0.2013} \\
    \hline
    \bottomrule
  \end{tabular}%
}
\begin{tablenotes}
\footnotesize
  \item $^{\dagger}$Higher values indicate better performance (mIoU, Dice). $^{\ddagger}$Lower values indicate better performance (XOR, HD). \textbf{Bold} values indicate best performance for each metric. 
\end{tablenotes}
\end{table}

To determine whether these performance improvements were statistically meaningful, paired \(t\)-tests were conducted between Crack-Segmenter and each baseline method, as shown in Table \ref{tab:statistical_tests}. The model exhibited statistically significant improvements in both mIoU and Dice score compared to all baseline methods, as indicated by positive mean differences and very low \(p\)-values (\(p < 0.01\) and mostly \(p < 0.001\)). For instance, when compared with commonly used segmentation models such as U-Net and DeepLabV3, the proposed method demonstrated substantial mean improvements in mIoU (+0.2396 and +0.2921, respectively) and Dice score (+0.1639 and +0.2144, respectively), all statistically significant at \(p < 0.01\).

\begin{figure}[H]
    \centering
    \includegraphics[width=1.0\textwidth]{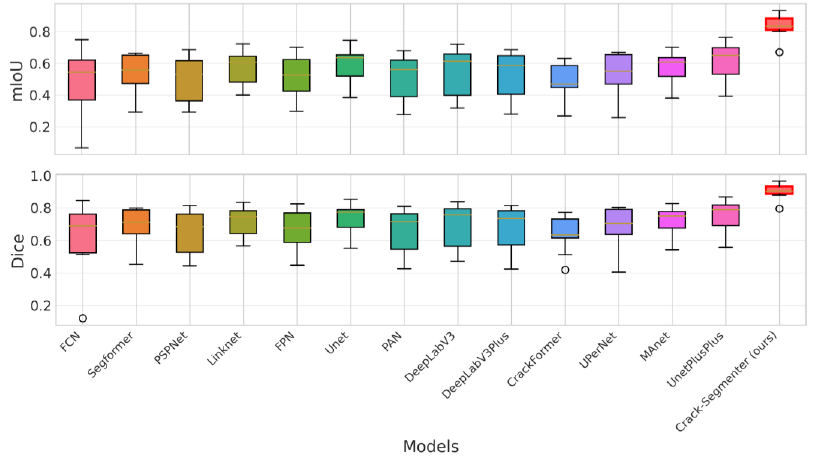}  % change file name and width as needed
    \caption{Box plot showing the mIoU and Dice Scores of Crack-Segmenter against other model baselines. }
    \label{fig:box plot}
\end{figure}

Furthermore, the highest statistical significance was noted against the CrackFormer model, with a mean difference of +0.3544 in mIoU and +0.2647 in Dice score (both \(p < 0.001\)). This strong statistical evidence confirms the effectiveness of integrating multi-scale embeddings, directional attention mechanisms, and adaptive fusion within Crack-Segmenter. These improvements likely stem from the enhanced capability of Crack-Segmenter to accurately capture diverse crack patterns and textures, as demonstrated consistently across datasets. Figure \ref{fig:box plot} shows distribution of the dice and mIoU scores of Crack-Segmenter and the baseline models. 

\begin{table*}[ht!]
\centering
\caption{Statistical significance tests comparing Crack‐Segmenter against baseline methods. Mean differences, t‐statistics, p‐values, and significance levels are reported for each evaluation metric. Positive mean differences for mIoU and Dice indicate superior performance.}
\label{tab:statistical_tests}

%--------------------------------------------------------------------
% Sub‐table 1: FCN, SegFormer, PSPNet, LinkNet, FPN, U-Net
%--------------------------------------------------------------------
\resizebox{\textwidth}{!}{%
\begin{tabular}{l l c c c c c c}
\toprule
% Heading for model columns
& & \multicolumn{6}{c}{\textbf{Models}} \\
\cmidrule(lr){3-8}
\textbf{Metric} & \textbf{Statistic} 
  & \textbf{FCN} & \textbf{SegFormer} & \textbf{PSPNet} 
  & \textbf{LinkNet} & \textbf{FPN} & \textbf{U-Net} \\
\midrule
\multirow{4}{*}{mIoU} 
  & mean diff.   & +0.3446 & +0.2968 & +0.3387 & +0.2581 & +0.3220 & +0.2396 \\
  & t-statistic  & 4.829   & 6.091   & 5.840   & 5.560   & 5.657   & 5.089   \\
  & p-value      & 0.000935& 0.000181& 0.000247& 0.000352& 0.000311& 0.000655\\
  & significance & ***     & ***     & ***     & ***     & ***     & ***     \\
\midrule
\multirow{4}{*}{Dice} 
  & mean Diff.   & +0.2779 & +0.2128 & +0.2562 & +0.1828 & +0.2415 & +0.1639 \\
  & t-statistic  & 3.886   & 5.204   & 5.084   & 5.125   & 4.927   & 4.586   \\
  & p-value      & 0.003696& 0.000561& 0.000659& 0.000624& 0.000816& 0.001317\\
  & significance & **      & ***     & ***     & ***     & ***     & **      \\
\bottomrule
\end{tabular}%
}

\vspace{1.2em}

%--------------------------------------------------------------------
% Sub‐table 2: PAN, DeepLabV3, DeepLabV3+, CrackFormer, UPerNet, MANet, U-Net++
%--------------------------------------------------------------------
\resizebox{\textwidth}{!}{%
\begin{tabular}{l l c c c c c c c}
% \toprule
% Heading for model columns
% & & \multicolumn{7}{c}{\textbf{Models}} \\
\cmidrule(lr){3-9}
\textbf{Metric} & \textbf{Statistic} 
  & \textbf{PAN} & \textbf{DeepLabV3} & \textbf{DeepLabV3+}
  & \textbf{CrackFormer} & \textbf{UPerNet} 
  & \textbf{MANet} & \textbf{U-Net++} \\
\midrule
\multirow{4}{*}{mIoU} 
  & mean diff.   & +0.3281 & +0.2921  & +0.3088  & +0.3544    & +0.3006 & +0.2618 & +0.2290 \\
  & t-statistic  & 5.682   & 4.852    & 5.394    & 7.523      & 5.837   & 5.649   & 4.449   \\
  & p-value      & 0.000301& 0.000906 & 0.000437 & 0.000036   & 0.000248& 0.000314& 0.001603\\
  & significance & ***     & ***      & ***      & ***        & ***     & ***     & **      \\
\midrule
\multirow{4}{*}{Dice} 
  & mean Diff.   & +0.2468 & +0.2144  & +0.2295  & +0.2647    & +0.2178 & +0.1843 & +0.1568 \\
  & t-statistic  & 4.799   & 4.180    & 4.573    & 6.516      & 4.887   & 4.993   & 3.968   \\
  & p-value      & 0.000975& 0.002377 & 0.001341 & 0.000109   & 0.000863& 0.000746& 0.003264\\
  & significance & ***     & **       & **       & ***        & ***     & ***     & **      \\
\bottomrule
\end{tabular}%
}

\begin{tablenotes}
  \small
  \item \textbf{Note:} *** \(p<0.001\), ** \(p<0.01\), * \(p<0.05\). Paired t‐tests assume the same dataset was used across all models.
\end{tablenotes}

\end{table*}

\subsection{Qualitative Results}
\begin{figure}[H]
    \centering    \includegraphics[width=1.0\textwidth]{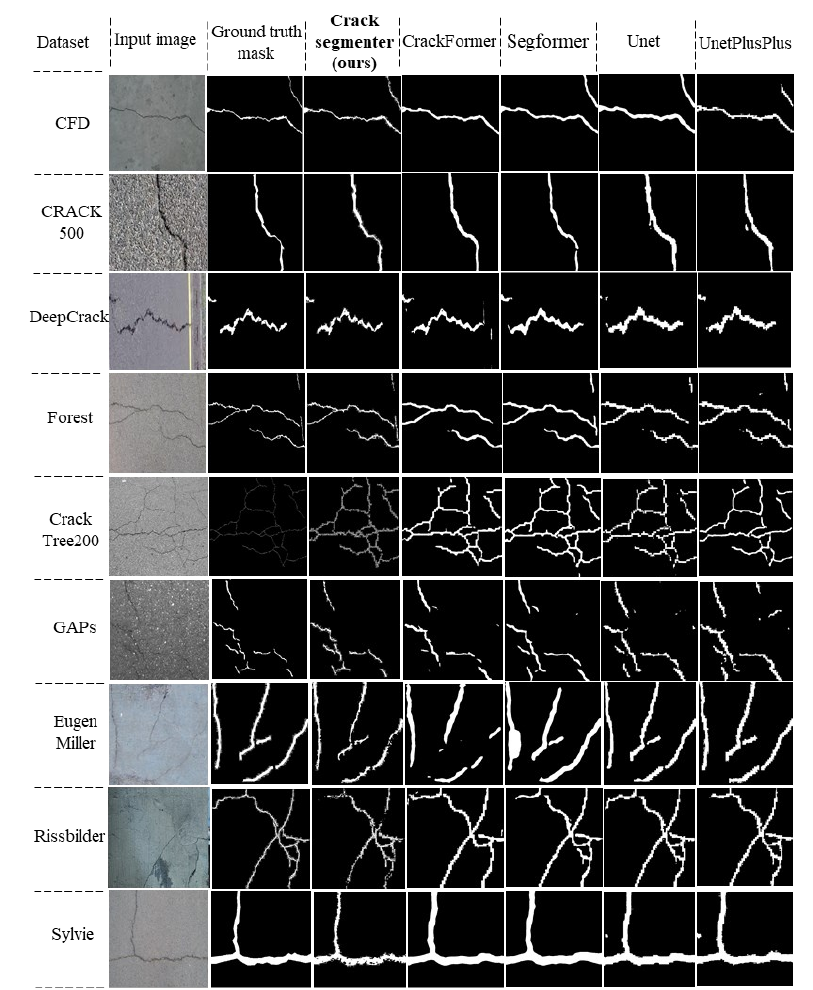}  % change file name and width as needed
    \caption{Qualitative comparison of predicted segmentation masks across multiple datasets. Each row shows the input image, ground truth, and predictions from the chosen baseline models (CrackFormer, Segformer, Unet, UnetPlusPlus) and the proposed \textit{Crack-Segmenter}. }
    \label{fig:qualitative results}
\end{figure}

Figure \ref{fig:qualitative results} presents visual comparisons between Crack-Segmenter and supervised baselines across nine datasets. Each row displays a different dataset, enabling comprehensive assessment of segmentation performance across diverse crack patterns and imaging conditions.

On the CFD dataset with thin linear cracks, the proposed method produces clean, continuous segmentations closely matching ground truth. While baseline methods capture general crack structure, they show inconsistent crack width. The multi-scale embeddings effectively capture fine details at appropriate resolutions, resulting in well-defined boundaries without excessive dilation.

For CRACK500's vertical cracks with varying widths, Crack-Segmenter maintains consistent segmentation quality throughout, accurately preserving narrow and wider sections. The Directional Attention Transformer emphasizes vertical continuity, leading to better boundary precision and endpoint detection compared to baselines.

The DeepCrack dataset contains horizontal cracks with irregular edges. All methods perform competently, but Crack-Segmenter shows superior noise suppression while maintaining sharp boundaries. The Attention-Guided Fusion module balances detail preservation with contextual consistency, reducing background artifacts present in baseline predictions.
On Forest's multiple branching cracks, Crack-Segmenter excels at maintaining connectivity at junctions, preserving complete network structure.

CrackTree200 presents dense interconnected crack networks. Crack-Segmenter captures intricate patterns while maintaining clear boundaries between closely spaced cracks. Multi-scale processing enables simultaneous detection of major branches and fine subsidiary cracks, whereas baselines either over-segment (merging adjacent cracks) or under-segment (missing finer branches).

For GAPs' diagonal cracks with variable contrast, Crack-Segmenter demonstrates robust performance even in low-contrast regions, maintaining consistent quality along entire crack lengths. The self-supervised learning identifies patterns based on structural characteristics rather than intensity contrasts alone.

The Eugen Miller dataset contains curved tunnel cracks under varying lighting. Crack-Segmenter provides smoother curve representations with fewer discretization artifacts than baselines. The directional attention adapts to changing crack orientations effectively.

On Rissbilder's intersecting cracks, Crack-Segmenter accurately segments both primary cracks and intersections without creating artificial connections or gaps. This demonstrates effective handling of complex spatial relationships through attention-guided fusion.

For Sylvie's prominent horizontal cracks, all methods achieve satisfactory results, but Crack-Segmenter produces the most consistent width representation along the entire length. This consistency stems from balanced feature extraction across scales.

These qualitative results confirm that Crack-Segmenter achieves segmentation quality matching or exceeding supervised methods while eliminating manual annotation requirements. The visual evidence demonstrates superior crack continuity preservation, consistent boundary maintenance, and adaptation to diverse crack morphologies across different pavement conditions. 

\begin{figure}[H]
    \centering    \includegraphics[width=0.7\textwidth]{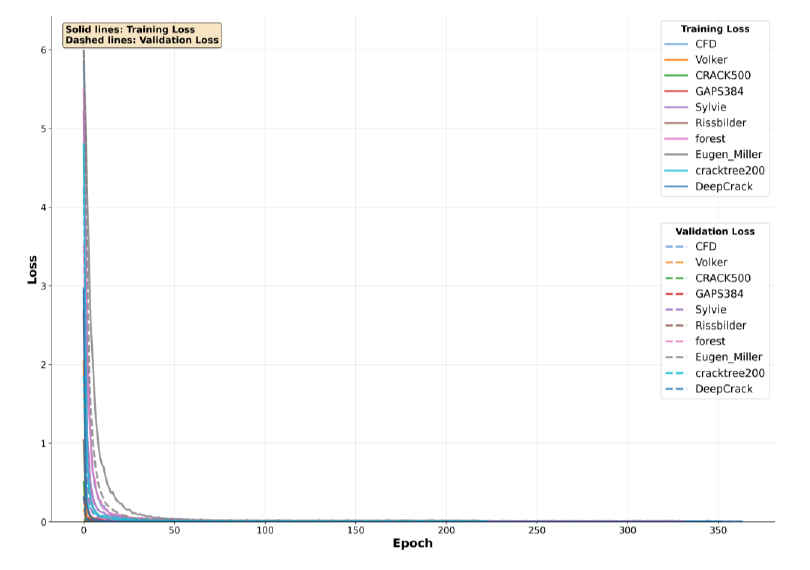}  % change file name and width as needed
    \caption{Training and Validation losses for Crack-Segmenter training accross all datasets}
    \label{fig:combined_loss}
\end{figure}

\begin{figure}[H]
    \centering    \includegraphics[width=0.7\textwidth]{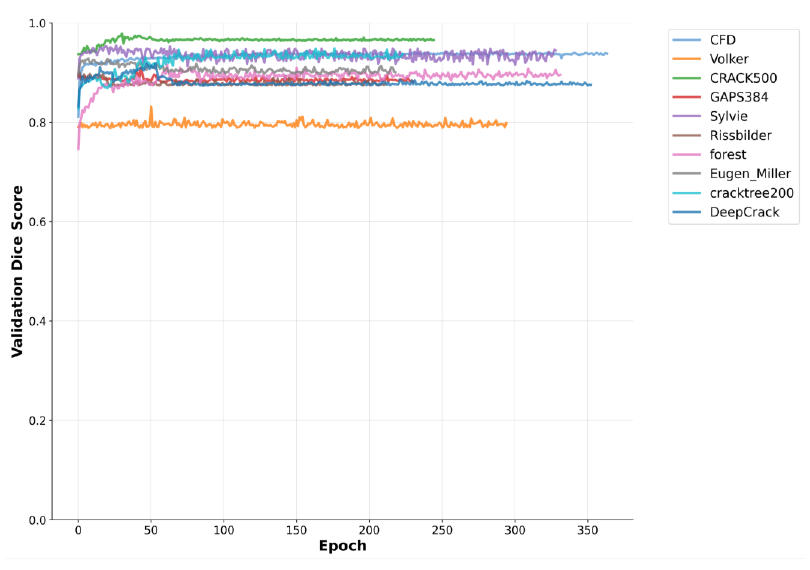}  % change file name and width as needed
    \caption{Validation Dice Scores for Cracksegmenter Training across all datasets}
    \label{fig:combined_dice}
\end{figure}

\begin{figure}[H]
    \centering    \includegraphics[width=0.7\textwidth]{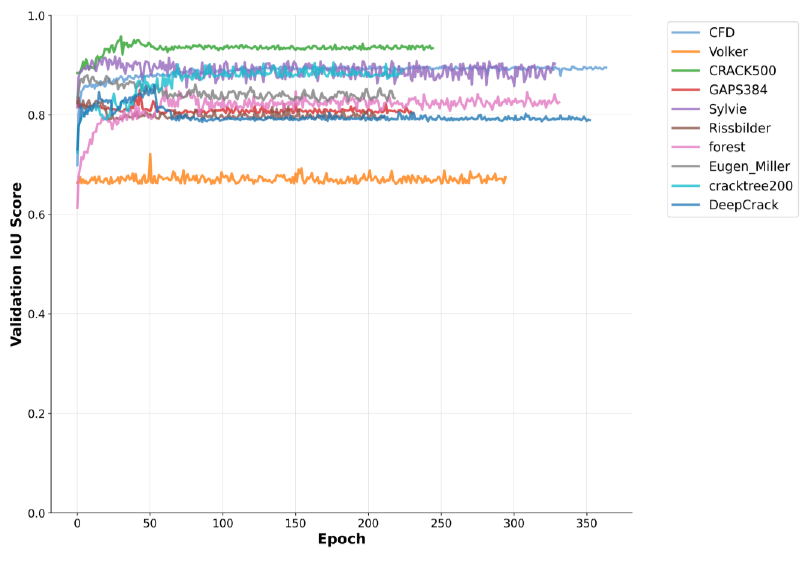}  % change file name and width as needed
    \caption{Validation IoU Scores for Cracksegmenter Training across all datasets}
    \label{fig:combined_iou}
\end{figure}

Figure~\ref{fig:combined_loss}, Figure~\ref{fig:combined_dice}, and Figure~\ref{fig:combined_iou} show the training loss, validation Dice score, and validation IoU convergence curves, respectively.

\subsection{Comparison with Unsupervised and Semi-Supervised Approaches}
To provide a broader context beyond fully supervised baselines, the study also compared Crack-Segmenter with representative unsupervised and semi-supervised crack segmentation methods on the DeepCrack dataset. These approaches are of particular interest because they aim to reduce or eliminate reliance on manual annotations, aligning with the motivation of the study. Table~\ref{tab:unsupervised_semi} summarizes the results.

\begin{table}[H]
    \centering
    \caption{Performance comparison of unsupervised, semi-supervised, and the proposed self-supervised Crack-Segmenter on the DeepCrack dataset.}
    \label{tab:unsupervised_semi}
    \begin{tabular}{lcc}
        \hline
        \textbf{Method} & \textbf{mIoU} & \textbf{Dice Score} \\
        \hline
        \multicolumn{3}{c}{\textit{Unsupervised Methods}} \\
        RIAD \cite{RIAD} & 0.2028 & 0.3372 \\
        SCADN \cite{SCADN} & 0.2919 & 0.4519 \\
        UP-CrackNet \cite{UP-CrackNet} & 0.5874 & 0.7401 \\
        \hline
        \multicolumn{3}{c}{\textit{Semi-Supervised Methods}} \\
        Mohammed et al. \cite{Mohammed} & 0.7360 & 0.8479 \\
        UniMatch \cite{Zhang2025DeepLF} & 0.6750 & 0.7930 \\
        UniMatchv2 \cite{Zhang2025DeepLF} & 0.7320 & 0.8400 \\
        \hline
        \multicolumn{3}{c}{\textit{Ours}} \\
        Crack-Segmenter & \textbf{0.8217} & \textbf{0.8952} \\
        \hline
    \end{tabular}
\end{table}

As shown in Table~\ref{tab:unsupervised_semi}, classical unsupervised approaches such as RIAD and SCADN perform poorly, achieving mIoU scores below 0.30 and Dice scores below 0.46, which demonstrates the difficulty of learning crack structures without supervision. UP-CrackNet achieves stronger performance (mIoU of 0.5874 and Dice score of 0.7401), but its reliance on reconstruction residuals introduces frequent false positives from background textures. Semi-supervised methods perform considerably better: Mohammed et al. \cite{Mohammed} reaches an mIoU of 0.7360 and a Dice score of 0.8479, while UniMatch and UniMatchv2 show competitive results by leveraging limited labeled data with consistency-based training.

In contrast, Crack-Segmenter consistently outperforms both unsupervised and semi-supervised methods, achieving an mIoU of 0.8217 and a Dice score of 0.8952. These results confirm that the proposed self-supervised framework is not only annotation-free but also achieves higher segmentation accuracy than existing paradigms that still rely on labels or indirect reconstruction strategies. This demonstrates the robustness and practical utility of the method for pavement crack segmentation under real-world conditions where annotated data are scarce or unavailable.

\section{Model Efficiency and Complexity}
To address computational efficiency concerns, the study conducted a comprehensive comparison of Crack-Segmenter against baseline segmentation models across key performance metrics. Table~\ref{tab:inference_comparison} presents the inference performance analysis on the DeepCrack dataset, evaluating parameter count, computational complexity, memory footprint, and inference speed.

Crack-Segmenter demonstrates exceptional parameter efficiency with only 3.37M parameters, representing a significant reduction compared to most baseline models. This parameter efficiency translates directly to memory efficiency, requiring only 12.85 MB of storage, substantially smaller than conventional architectures such as UNet (124.27 MB) and UNetPlusPlus (187.10 MB). However, its computational complexity is higher, requiring 142.1 GFLOPs compared to more efficient baselines like PSPNet (9.1 GFLOPs) and SegFormer (23.0 GFLOPs).

The computational demands impact inference speed, with a latency of 199.03 ms compared to faster alternatives such as PSPNet (3.18 ms) and FPN (6.82 ms). This latency translates to 5.02 FPS, reflecting the computational cost of the architecture design. While the proposed approach requires more inference time than CrackFormer (67.40 ms), this computational cost is justified by the superior segmentation performance as demonstrated in previous sections.

The efficiency analysis reveals a deliberate trade-off in Crack-Segmenter's architecture design between computational complexity and model compactness. While the proposed method requires more FLOPs during inference, it achieves this performance with significantly fewer parameters and minimal memory footprint, making it suitable for deployment scenarios where model size and memory constraints are critical considerations.

\begin{table}[H]
\centering
\caption{Inference Performance Comparison of Crack-Segmenter and Baseline models on DeepCrack Dataset}
\label{tab:inference_comparison}
\resizebox{\textwidth}{!}{%
\begin{tabular}{lccccc}
\toprule
\textbf{Model} & \textbf{Params (M)} & \textbf{FLOPs (G)} & \textbf{Model Size (MB)} & \textbf{Latency (ms)} & \textbf{FPS} \\
\midrule
Unet & 32.52 & 32.8 & 124.27 & 9.98 & 100.18 \\
UNetPlusPlus & 48.99 & 176.3 & 187.10 & 25.96 & 38.51 \\
PSPNet & 24.30 & \textbf{9.1} & 92.91 & \textbf{3.18} & \textbf{314.72} \\
PAN & 24.26 & 26.7 & 92.75 & 7.24 & 138.18 \\
MANet & 147.44 & 57.1 & 562.68 & 13.36 & 74.85 \\
LinkNet & 31.18 & 33.0 & 119.17 & 8.09 & 123.57 \\
FPN & 26.12 & 24.0 & 99.83 & 6.82 & 146.54 \\
DeepLabV3 & 39.63 & 125.7 & 151.41 & 15.55 & 64.32 \\
DeepLabV3Plus & 26.68 & 28.3 & 101.99 & 7.21 & 138.61 \\
UPerNet & 37.28 & 58.7 & 142.46 & 10.04 & 99.65 \\
SegFormer & 24.84 & 23.0 & 94.95 & 8.21 & 121.87 \\
CrackFormer & 4.96 & 62.7 & 18.93 & 67.40 & 14.84 \\
\midrule
\textbf{Crack-Segmenter (Ours)} & \textbf{3.37} & 142.1 & \textbf{12.85} & 199.03 & 5.02 \\
\bottomrule
\end{tabular}%
}
\end{table}

\section{Practical Applications and Limitations}
The proposed \textit{Crack-Segmenter} framework offers significant practical benefits in construction and infrastructure management, where efficient pavement crack detection is vital for ensuring road safety, enhancing infrastructure durability, and optimizing asset management. Eliminating manual annotation costs enables transportation agencies to conduct regular, comprehensive inspections more affordably, supporting preventive pavement maintenance planning through accurate crack segmentation maps that facilitate early identification and timely repair of pavement defects. The approach seamlessly integrates with automated inspection technologies, such as unmanned aerial vehicles (UAVs) and vehicle-mounted systems, enabling rapid, continuous monitoring of large road networks with minimal human involvement. This automation particularly benefits construction and engineering firms for proactive monitoring of new or rehabilitated pavement surfaces, allowing swift identification of potential structural weaknesses and timely corrective measures. The annotation-free nature further democratizes access to advanced pavement assessment tools in resource-constrained regions, supporting equitable infrastructure management practices globally.

However, several limitations may affect broader applicability and deployment scenarios. The proposed method incurs relatively high computational cost during inference due to its multi-scale transformer architecture and attention-guided fusion, requiring more floating-point operations and having higher latency than lightweight alternatives. This computational overhead presents challenges for deployment in real-time or low-power environments, such as UAV-based inspections or mobile edge devices, potentially limiting the automated inspection capabilities described above. Additionally, the model exhibits varying performance across datasets, particularly achieving lower mIoU scores on the Volker dataset (0.6707) compared to U-NetPlusPlus (0.7641), suggesting limited generalizability in certain imaging conditions or surface types with subtle visual distinctions. Furthermore, Crack-Segmenter focuses exclusively on binary crack segmentation without distinguishing between different crack types or accounting for other pavement distresses such as potholes, rutting, or patching. Despite these limitations, the proposed architecture serves as a promising foundation for scalable, annotation-free crack detection, with future work addressing these challenges through architectural optimization, domain adaptation, and multi-label learning to enhance practical impact.

\section{Conclusion}
This paper presented \textit{Crack-Segmenter}, a fully self-supervised segmentation framework developed specifically for pavement crack detection. Crack-Segmenter successfully eliminates the dependence on costly and labor-intensive pixel-level annotations, addressing a significant limitation of existing segmentation methods. The proposed model integrates three innovative modules: the Scale-Adaptive Embedder, Directional Attention Transformer, and Attention-Guided Fusion. Each module targets specific challenges in crack segmentation, collectively enhancing the model’s capability to accurately detect diverse crack patterns.

Experimental evaluations conducted on ten publicly available datasets demonstrated the effectiveness  of the proposed framework. Crack-Segmenter significantly outperformed state-of-the-art fully supervised methods across multiple metrics, including mIoU, Dice score, XOR, and HD metric. Comprehensive statistical analysis confirmed the statistical significance of these improvements, with notably strong performance gains observed over prominent baseline methods.

Ablation studies further underscored the importance of each proposed module. Specifically, SAE facilitated effective multi-scale feature extraction, capturing cracks across varied widths and complexities. DAT enhanced spatial coherence and continuity, crucial for linear crack structures. Finally, AGF intelligently fused these features, emphasizing contextually relevant information at each scale. Collectively, these modules delivered superior segmentation accuracy, confirming their individual and combined effectiveness.

Attention map visualizations provided interpretability, confirming that Crack-Segmenter correctly focused on meaningful crack regions, reducing background distractions. Such explainability strengthens confidence in deploying this method in practical, real-world pavement monitoring tasks.

In future work, extending this framework to handle other pavement distress types and further optimizing computational efficiency could expand its utility. Additionally, exploring methods to integrate temporal information from sequential pavement inspections may improve detection robustness over time, enhancing preventive maintenance practices.

\section{Data Availability}
The datasets used in this paper are all publicly available. Also, source code and training scripts for Crack-Segmenter are publicly available at \url{\https://github.com/Blessing988/Crack-Segmenter}.

\section{Acknowledgemnts}
This research was funded by the Upper Great Plains Transportation Institute (UGPTI) at North Dakota State University under the University Transportation Center (UTC) Grant, Project Number CTIPS-51.

\bibliographystyle{elsarticle-num}
\bibliography{export}

\begin{thebibliography}{10}
\expandafter\ifx\csname url\endcsname\relax
  \def\url#1{\texttt{#1}}\fi
\expandafter\ifx\csname urlprefix\endcsname\relax\def\urlprefix{URL }\fi
\expandafter\ifx\csname href\endcsname\relax
  \def\href#1#2{#2} \def\path#1{#1}\fi

\bibitem{Kyem2024PaveCapTF}
B.~A. Kyem, E.~K.~O. Denteh, J.~K. Asamoah, A.~Aboah, \href{https://api.semanticscholar.org/CorpusID:271768959}{Pavecap: The first multimodal framework for comprehensive pavement condition assessment with dense captioning and pci estimation}, ArXiv abs/2408.04110 (2024).
\newline\urlprefix\url{https://api.semanticscholar.org/CorpusID:271768959}

\bibitem{barman2019cracksealing}
M.~Barman, J.~Munch, U.~M. Arepalli, \href{https://rosap.ntl.bts.gov/view/dot/61830}{Cost/benefit analysis of the effectiveness of crack sealing techniques}, Research Report MN/RC 2019-26, University of Minnesota, Duluth. Department of Civil Engineering, prepared for the Local Road Research Board (June 2019).
\newline\urlprefix\url{https://rosap.ntl.bts.gov/view/dot/61830}

\bibitem{BlessingAIC25}
B.~A. Kyem, J.~N. Owor, A.~Danyo, J.~K. Asamoah, E.~Denteh, T.~Muturi, A.~Dontoh, Y.~Adu-Gyamfi, A.~Aboah, Task-specific dual-model framework for comprehensive traffic safety video description and analysis, in: The IEEE International Conference on Computer Vision (ICCV) Workshops, 2025.

\bibitem{claim_ssl}
M.~Sohaib, M.~J. Hasan, M.~Shah, Z.~Zheng, \href{https://www.nature.com/articles/s41598-024-63575-x}{A robust self-supervised approach for fine-grained crack detection in concrete structures}, Scientific Reports 14 (06 2024).
\newblock \href {https://doi.org/10.1038/s41598-024-63575-x} {\path{doi:10.1038/s41598-024-63575-x}}.
\newline\urlprefix\url{https://www.nature.com/articles/s41598-024-63575-x}

\bibitem{gans_ssl}
K.~Zhang, Y.~Zhang, H.~Cheng, \href{https://ascelibrary.org/doi/abs/10.1061/(ASCE)CP.1943-5487.0000883}{Self-supervised structure learning for crack detection based on cycle-consistent generative adversarial networks}, Journal of Computing in Civil Engineering 34 (2020) 04020004.
\newblock \href {https://doi.org/10.1061/(ASCE)CP.1943-5487.0000883} {\path{doi:10.1061/(ASCE)CP.1943-5487.0000883}}.
\newline\urlprefix\url{https://ascelibrary.org/doi/abs/10.1061/(ASCE)CP.1943-5487.0000883}

\bibitem{Lau2020Automated}
S.~L.~H. Lau, E.~Chong, X.~Yang, X.~Wang, \href{https://ieeexplore.ieee.org/document/9121269}{Automated pavement crack segmentation using u-net-based convolutional neural network}, IEEE Access 8 (2020) 114892--114899.
\newblock \href {https://doi.org/10.1109/ACCESS.2020.3003638} {\path{doi:10.1109/ACCESS.2020.3003638}}.
\newline\urlprefix\url{https://ieeexplore.ieee.org/document/9121269}

\bibitem{kyem_context_cracknet}
B.~{Agyei Kyem}, J.~K. Asamoah, A.~Aboah, \href{https://www.sciencedirect.com/science/article/pii/S0950061825017337}{Context-cracknet: A context-aware framework for precise segmentation of tiny cracks in pavement images}, Construction and Building Materials 484 (2025) 141583.
\newblock \href {https://doi.org/https://doi.org/10.1016/j.conbuildmat.2025.141583} {\path{doi:https://doi.org/10.1016/j.conbuildmat.2025.141583}}.
\newline\urlprefix\url{https://www.sciencedirect.com/science/article/pii/S0950061825017337}

\bibitem{PDSNET}
T.~Wen, S.~Ding, H.~Lang, J.~J. Lu, Y.~Yuan, Y.~Peng, J.~Chen, A.~Wang, \href{https://doi.org/10.1080/10298436.2022.2027414}{Automated pavement distress segmentation on asphalt surfaces using a deep learning network}, International Journal of Pavement Engineering 24~(2) (2023) 2027414.
\newblock \href {https://doi.org/10.1080/10298436.2022.2027414} {\path{doi:10.1080/10298436.2022.2027414}}.
\newline\urlprefix\url{https://doi.org/10.1080/10298436.2022.2027414}

\bibitem{DUNG201952}
C.~V. Dung, L.~D. Anh, \href{https://www.sciencedirect.com/science/article/pii/S0926580518306745}{Autonomous concrete crack detection using deep fully convolutional neural network}, Automation in Construction 99 (2019) 52--58.
\newblock \href {https://doi.org/https://doi.org/10.1016/j.autcon.2018.11.028} {\path{doi:https://doi.org/10.1016/j.autcon.2018.11.028}}.
\newline\urlprefix\url{https://www.sciencedirect.com/science/article/pii/S0926580518306745}

\bibitem{Yang_fpn}
F.~Yang, L.~Zhang, S.~Yu, D.~Prokhorov, X.~Mei, H.~Ling, \href{https://ieeexplore.ieee.org/document/8694955}{Feature pyramid and hierarchical boosting network for pavement crack detection}, IEEE Transactions on Intelligent Transportation Systems 21~(4) (2020) 1525--1535.
\newblock \href {https://doi.org/10.1109/TITS.2019.2910595} {\path{doi:10.1109/TITS.2019.2910595}}.
\newline\urlprefix\url{https://ieeexplore.ieee.org/document/8694955}

\bibitem{deepcrack}
Y.~Liu, J.~Yao, X.~Lu, R.~Xie, L.~Li, \href{https://www.sciencedirect.com/science/article/pii/S0925231219300566}{Deepcrack: A deep hierarchical feature learning architecture for crack segmentation}, Neurocomputing 338 (2019) 139--153.
\newblock \href {https://doi.org/https://doi.org/10.1016/j.neucom.2019.01.036} {\path{doi:https://doi.org/10.1016/j.neucom.2019.01.036}}.
\newline\urlprefix\url{https://www.sciencedirect.com/science/article/pii/S0925231219300566}

\bibitem{jenkins2018deep}
M.~D. Jenkins, T.~A. Carr, M.~I. Iglesias, T.~Buggy, G.~Morison, \href{https://ieeexplore.ieee.org/document/8694955}{A deep convolutional neural network for semantic pixel-wise segmentation of road and pavement surface cracks}, in: 2018 26th European signal processing conference (EUSIPCO), IEEE, 2018, pp. 2120--2124.
\newline\urlprefix\url{https://ieeexplore.ieee.org/document/8694955}

\bibitem{PAN2020103357}
Y.~Pan, G.~Zhang, L.~Zhang, \href{https://www.sciencedirect.com/science/article/pii/S0926580520309377}{A spatial-channel hierarchical deep learning network for pixel-level automated crack detection}, Automation in Construction 119 (2020) 103357.
\newblock \href {https://doi.org/https://doi.org/10.1016/j.autcon.2020.103357} {\path{doi:https://doi.org/10.1016/j.autcon.2020.103357}}.
\newline\urlprefix\url{https://www.sciencedirect.com/science/article/pii/S0926580520309377}

\bibitem{GUO2023104646}
F.~Guo, Y.~Qian, J.~Liu, H.~Yu, \href{https://www.sciencedirect.com/science/article/pii/S0926580522005167}{Pavement crack detection based on transformer network}, Automation in Construction 145 (2023) 104646.
\newblock \href {https://doi.org/https://doi.org/10.1016/j.autcon.2022.104646} {\path{doi:https://doi.org/10.1016/j.autcon.2022.104646}}.
\newline\urlprefix\url{https://www.sciencedirect.com/science/article/pii/S0926580522005167}

\bibitem{CrackFormer}
H.~Liu, J.~Yang, X.~Miao, C.~Mertz, H.~Kong, \href{https://ieeexplore.ieee.org/document/9711107}{Crackformer network for pavement crack segmentation}, IEEE Transactions on Intelligent Transportation Systems 24~(9) (2023) 9240--9252.
\newblock \href {https://doi.org/10.1109/TITS.2023.3266776} {\path{doi:10.1109/TITS.2023.3266776}}.
\newline\urlprefix\url{https://ieeexplore.ieee.org/document/9711107}

\bibitem{Kyem2024AdvancingPD}
B.~A. Kyem, E.~K.~O. Denteh, J.~K. Asamoah, K.~A. Tutu, A.~Aboah, \href{https://api.semanticscholar.org/CorpusID:271854773}{Advancing pavement distress detection in developing countries: A novel deep learning approach with locally-collected datasets}, ArXiv abs/2408.05649 (2024).
\newline\urlprefix\url{https://api.semanticscholar.org/CorpusID:271854773}

\bibitem{ASAMOAH2025128003}
J.~K. Asamoah, B.~{Agyei Kyem}, N.~D. Obeng-Amoako, A.~Aboah, \href{https://www.sciencedirect.com/science/article/pii/S0957417425016240}{Saam-reflectnet: Sign-aware attention-based multitasking framework for integrated traffic sign detection and retroreflectivity estimation}, Expert Systems with Applications 286 (2025) 128003.
\newblock \href {https://doi.org/https://doi.org/10.1016/j.eswa.2025.128003} {\path{doi:https://doi.org/10.1016/j.eswa.2025.128003}}.
\newline\urlprefix\url{https://www.sciencedirect.com/science/article/pii/S0957417425016240}

\bibitem{XIANG2024108497}
C.~Xiang, V.~J. Gan, L.~Deng, J.~Guo, S.~Xu, \href{https://www.sciencedirect.com/science/article/pii/S0952197624006559}{Unified weakly and semi-supervised crack segmentation framework using limited coarse labels}, Engineering Applications of Artificial Intelligence 133 (2024) 108497.
\newblock \href {https://doi.org/https://doi.org/10.1016/j.engappai.2024.108497} {\path{doi:https://doi.org/10.1016/j.engappai.2024.108497}}.
\newline\urlprefix\url{https://www.sciencedirect.com/science/article/pii/S0952197624006559}

\bibitem{LI2025105899}
J.~Li, C.~Yuan, X.~Wang, G.~Chen, G.~Ma, \href{https://www.sciencedirect.com/science/article/pii/S0926580524006356}{Semi-supervised crack detection using segment anything model and deep transfer learning}, Automation in Construction 170 (2025) 105899.
\newblock \href {https://doi.org/https://doi.org/10.1016/j.autcon.2024.105899} {\path{doi:https://doi.org/10.1016/j.autcon.2024.105899}}.
\newline\urlprefix\url{https://www.sciencedirect.com/science/article/pii/S0926580524006356}

\bibitem{pavesam}
A.~A. Neema Jakisa~Owor, Yaw Adu-Gyamfi, M.~Amo-Boateng, \href{https://doi.org/10.1080/14680629.2024.2374863}{Pavesam – segment anything for pavement distress}, Road Materials and Pavement Design 0~(0) (2024) 1--25.
\newblock \href {https://doi.org/10.1080/14680629.2024.2374863} {\path{doi:10.1080/14680629.2024.2374863}}.
\newline\urlprefix\url{https://doi.org/10.1080/14680629.2024.2374863}

\bibitem{MuturiAIC25}
T.~Muturi, B.~A. Kyem, J.~K. Asamoah, J.~N. Owor, R.~Dyzinela, A.~Danyo, Y.~Adu-Gyamfi, A.~Aboah, Prompt-guided spatial understanding with rgb-d transformers for fine-grained object relation reasoning, in: The IEEE International Conference on Computer Vision (ICCV) Workshops, 2025.

\bibitem{Li2020Semi-Supervised}
G.~Li, J.~Wan, S.~He, Q.~Liu, B.~Ma, \href{https://ieeexplore.ieee.org/document/9032091}{Semi-supervised semantic segmentation using adversarial learning for pavement crack detection}, IEEE Access 8 (2020) 51446--51459.
\newblock \href {https://doi.org/10.1109/ACCESS.2020.2980086} {\path{doi:10.1109/ACCESS.2020.2980086}}.
\newline\urlprefix\url{https://ieeexplore.ieee.org/document/9032091}

\bibitem{adversarial_gans}
S.~Shim, J.~Kim, G.-C. Cho, S.-W. Lee, \href{https://ieeexplore.ieee.org/document/9193940}{Multiscale and adversarial learning-based semi-supervised semantic segmentation approach for crack detection in concrete structures}, IEEE Access 8 (2020) 170939--170950.
\newblock \href {https://doi.org/10.1109/ACCESS.2020.3022786} {\path{doi:10.1109/ACCESS.2020.3022786}}.
\newline\urlprefix\url{https://ieeexplore.ieee.org/document/9193940}

\bibitem{stargan}
B.~A. Kyem, J.~K. Asamoah, Y.~Huang, A.~Aboah, \href{https://ieeexplore.ieee.org/document/10807178}{Weather-adaptive synthetic data generation for enhanced power line inspection using stargan}, IEEE Access 12 (2024) 193882--193901.
\newblock \href {https://doi.org/10.1109/ACCESS.2024.3520120} {\path{doi:10.1109/ACCESS.2024.3520120}}.
\newline\urlprefix\url{https://ieeexplore.ieee.org/document/10807178}

\bibitem{Wang2021Semi-supervised}
W.~Wang, C.~Su, Semi-supervised semantic segmentation network for surface crack detection, Automation in Construction (2021).
\newblock \href {https://doi.org/10.1016/J.AUTCON.2021.103786} {\path{doi:10.1016/J.AUTCON.2021.103786}}.

\bibitem{triplet_teacher}
M.~Mohammed, Z.~Han, Y.~Li, Z.~Al-Huda, W.-D. Wang, \href{https://www.tandfonline.com/doi/full/10.1080/10298436.2024.2400562}{Enhanced pavement crack segmentation with minimal labeled data: a triplet attention teacher-student framework}, International Journal of Pavement Engineering 2024, VOL. 25 (2024) 2400562.
\newblock \href {https://doi.org/10.1080/10298436.2024.2400562} {\path{doi:10.1080/10298436.2024.2400562}}.
\newline\urlprefix\url{https://www.tandfonline.com/doi/full/10.1080/10298436.2024.2400562}

\bibitem{teacher_pseudo}
Z.~Jian, J.~Liu, \href{https://doi.org/10.1016/j.aei.2023.102279}{Cross teacher pseudo supervision: Enhancing semi-supervised crack segmentation with consistency learning}, Adv. Eng. Inform. 59~(C) (Jan. 2024).
\newblock \href {https://doi.org/10.1016/j.aei.2023.102279} {\path{doi:10.1016/j.aei.2023.102279}}.
\newline\urlprefix\url{https://doi.org/10.1016/j.aei.2023.102279}

\bibitem{Feng2021GCN-Based}
H.~Feng, W.~Li, Z.~Luo, Y.~Chen, S.~Fatholahi, M.~Cheng, C.~Wang, J.~M. Junior, J.~Li, \href{https://ieeexplore.ieee.org/document/9508901}{Gcn-based pavement crack detection using mobile lidar point clouds}, IEEE Transactions on Intelligent Transportation Systems 23 (2021) 11052--11061.
\newblock \href {https://doi.org/10.1109/tits.2021.3099023} {\path{doi:10.1109/tits.2021.3099023}}.
\newline\urlprefix\url{https://ieeexplore.ieee.org/document/9508901}

\bibitem{contrastive}
H.~Feng, L.~Ma, Y.~Yu, Y.~Chen, J.~Li, \href{https://www.sciencedirect.com/science/article/pii/S1569843223000705}{Scl-gcn: Stratified contrastive learning graph convolution network for pavement crack detection from mobile lidar point clouds}, International Journal of Applied Earth Observation and Geoinformation 118 (2023) 103248.
\newblock \href {https://doi.org/https://doi.org/10.1016/j.jag.2023.103248} {\path{doi:https://doi.org/10.1016/j.jag.2023.103248}}.
\newline\urlprefix\url{https://www.sciencedirect.com/science/article/pii/S1569843223000705}

\bibitem{Denteh2025IntegratingTB}
E.~K.~O. Denteh, A.~Danyo, J.~K. Asamoah, B.~A. Kyem, T.~Addai, A.~Aboah, \href{https://api.semanticscholar.org/CorpusID:277349514}{Integrating travel behavior forecasting and generative modeling for predicting future urban mobility and spatial transformations}, ArXiv abs/2503.21158 (2025).
\newline\urlprefix\url{https://api.semanticscholar.org/CorpusID:277349514}

\bibitem{Denteh2025DemographicsInformedNN}
E.~K.~O. Denteh, A.~Danyo, J.~K. Asamoah, B.~A. Kyem, A.~Aboah, \href{https://api.semanticscholar.org/CorpusID:279402772}{Demographics-informed neural network for multi-modal spatiotemporal forecasting of urban growth and travel patterns using satellite imagery}, ArXiv abs/2506.12456 (2025).
\newline\urlprefix\url{https://api.semanticscholar.org/CorpusID:279402772}

\bibitem{Al-Huda2023}
Z.~Al-Huda, B.~Peng, R.~N.~A. Algburi, S.~Alfasly, T.~Li, \href{https://doi.org/10.1007/s10489-022-04212-w}{Weakly supervised pavement crack semantic segmentation based on multi-scale object localization and incremental annotation refinement}, Applied Intelligence 53~(11) (2023) 14527--14546.
\newblock \href {https://doi.org/10.1007/s10489-022-04212-w} {\path{doi:10.1007/s10489-022-04212-w}}.
\newline\urlprefix\url{https://doi.org/10.1007/s10489-022-04212-w}

\bibitem{HE2024134668}
T.~He, H.~Li, Z.~Qian, C.~Niu, R.~Huang, \href{https://www.sciencedirect.com/science/article/pii/S0950061823043891}{Research on weakly supervised pavement crack segmentation based on defect location by generative adversarial network and target re‐optimization}, Construction and Building Materials 411 (2024) 134668.
\newblock \href {https://doi.org/https://doi.org/10.1016/j.conbuildmat.2023.134668} {\path{doi:https://doi.org/10.1016/j.conbuildmat.2023.134668}}.
\newline\urlprefix\url{https://www.sciencedirect.com/science/article/pii/S0950061823043891}

\bibitem{Shim}
S.~Shim, J.~Kim, G.-C. Cho, S.-W. Lee, \href{https://ieeexplore.ieee.org/abstract/document/9193940}{Multiscale and adversarial learning-based semi-supervised semantic segmentation approach for crack detection in concrete structures}, IEEE Access 8 (2020) 170939--170950.
\newblock \href {https://doi.org/10.1109/ACCESS.2020.3022786} {\path{doi:10.1109/ACCESS.2020.3022786}}.
\newline\urlprefix\url{https://ieeexplore.ieee.org/abstract/document/9193940}

\bibitem{SHI2025109683}
T.~Shi, Y.~Wang, Y.~Fang, Y.~Zhang, \href{https://www.sciencedirect.com/science/article/pii/S0952197624018414}{Semi-supervised segmentation model for crack detection based on mutual consistency constraint and boundary loss}, Engineering Applications of Artificial Intelligence 139 (2025) 109683.
\newblock \href {https://doi.org/https://doi.org/10.1016/j.engappai.2024.109683} {\path{doi:https://doi.org/10.1016/j.engappai.2024.109683}}.
\newline\urlprefix\url{https://www.sciencedirect.com/science/article/pii/S0952197624018414}

\bibitem{HAN2024105332}
C.~Han, H.~Yang, T.~Ma, S.~Wang, C.~Zhao, Y.~Yang, \href{https://www.sciencedirect.com/science/article/pii/S0926580524000682}{Crackdiffusion: A two-stage semantic segmentation framework for pavement crack combining unsupervised and supervised processes}, Automation in Construction 160 (2024) 105332.
\newblock \href {https://doi.org/https://doi.org/10.1016/j.autcon.2024.105332} {\path{doi:https://doi.org/10.1016/j.autcon.2024.105332}}.
\newline\urlprefix\url{https://www.sciencedirect.com/science/article/pii/S0926580524000682}

\bibitem{OworAIC25}
N.~J. Owor, J.~K. Asamoah, T.~Muturi, J.~A. Owor, B.~A. Kyem, A.~Danyo, Y.~Adu-Gyamfi, A.~Aboah, A unified detection pipeline for robust object detection in fisheye-based traffic surveillance, in: The IEEE International Conference on Computer Vision (ICCV) Workshops, 2025.

\bibitem{Gui2023ASO}
J.~Gui, T.~Chen, J.~Zhang, Q.~Cao, Z.~Sun, H.~Luo, D.~Tao, \href{https://api.semanticscholar.org/CorpusID:261046875}{A survey on self-supervised learning: Algorithms, applications, and future trends}, IEEE Transactions on Pattern Analysis and Machine Intelligence 46 (2023) 9052--9071.
\newline\urlprefix\url{https://api.semanticscholar.org/CorpusID:261046875}

\bibitem{Caron2018DeepCF}
M.~Caron, P.~Bojanowski, A.~Joulin, M.~Douze, \href{https://api.semanticscholar.org/CorpusID:263891125}{Deep clustering for unsupervised learning of visual features}, in: European Conference on Computer Vision, 2018.
\newline\urlprefix\url{https://api.semanticscholar.org/CorpusID:263891125}

\bibitem{balestriero2023cookbookselfsupervisedlearning}
R.~Balestriero, M.~Ibrahim, V.~Sobal, A.~Morcos, S.~Shekhar, T.~Goldstein, F.~Bordes, A.~Bardes, G.~Mialon, Y.~Tian, A.~Schwarzschild, A.~G. Wilson, J.~Geiping, Q.~Garrido, P.~Fernandez, A.~Bar, H.~Pirsiavash, Y.~LeCun, M.~Goldblum, \href{https://arxiv.org/abs/2304.12210}{A cookbook of self-supervised learning} (2023).
\newline\urlprefix\url{https://arxiv.org/abs/2304.12210}

\bibitem{BEYENE2023107889}
D.~A. Beyene, D.~Q. Tran, M.~B. Maru, T.~Kim, S.~Park, S.~Park, \href{https://www.sciencedirect.com/science/article/pii/S2352710223020697}{Unsupervised domain adaptation-based crack segmentation using transformer network}, Journal of Building Engineering 80 (2023) 107889.
\newblock \href {https://doi.org/https://doi.org/10.1016/j.jobe.2023.107889} {\path{doi:https://doi.org/10.1016/j.jobe.2023.107889}}.
\newline\urlprefix\url{https://www.sciencedirect.com/science/article/pii/S2352710223020697}

\bibitem{NGUYEN2025105892}
Q.~D. Nguyen, H.-T. Thai, S.~D. Nguyen, \href{https://www.sciencedirect.com/science/article/pii/S0926580524006289}{Self-training method for structural crack detection using image blending-based domain mixing and mutual learning}, Automation in Construction 170 (2025) 105892.
\newblock \href {https://doi.org/https://doi.org/10.1016/j.autcon.2024.105892} {\path{doi:https://doi.org/10.1016/j.autcon.2024.105892}}.
\newline\urlprefix\url{https://www.sciencedirect.com/science/article/pii/S0926580524006289}

\bibitem{LIN2022104544}
Z.~Lin, H.~Wang, S.~Li, \href{https://www.sciencedirect.com/science/article/pii/S0926580522004150}{Pavement anomaly detection based on transformer and self-supervised learning}, Automation in Construction 143 (2022) 104544.
\newblock \href {https://doi.org/https://doi.org/10.1016/j.autcon.2022.104544} {\path{doi:https://doi.org/10.1016/j.autcon.2022.104544}}.
\newline\urlprefix\url{https://www.sciencedirect.com/science/article/pii/S0926580522004150}

\bibitem{anomaly_pave}
\href{https://doi.org/10.1115/DETC2019-98135}{A Self-Supervised Learning Technique for Road Defects Detection Based on Monocular Three-Dimensional Reconstruction}, Vol. Volume 3: 21st International Conference on Advanced Vehicle Technologies; 16th International Conference on Design Education of International Design Engineering Technical Conferences and Computers and Information in Engineering Conference.
\newblock \href {https://doi.org/10.1115/DETC2019-98135} {\path{doi:10.1115/DETC2019-98135}}.
\newline\urlprefix\url{https://doi.org/10.1115/DETC2019-98135}

\bibitem{sidewalk}
H.-Y. Yoon, J.-H. Kim, J.-W. Jeong, \href{https://www.mdpi.com/1424-8220/22/1/380}{Classification of the sidewalk condition using self-supervised transfer learning for wheelchair safety driving}, Sensors 22~(1) (2022).
\newblock \href {https://doi.org/10.3390/s22010380} {\path{doi:10.3390/s22010380}}.
\newline\urlprefix\url{https://www.mdpi.com/1424-8220/22/1/380}

\bibitem{two_stage_unet}
Q.~Song, W.~Yao, H.~Tian, Y.~Guo, R.~C. Muniyandi, Y.~An, \href{https://doi.org/10.1016/j.eswa.2023.122406}{Two-stage framework with improved u-net based on self-supervised contrastive learning for pavement crack segmentation}, Expert Syst. Appl. 238~(Part F) (2024) 122406.
\newline\urlprefix\url{https://doi.org/10.1016/j.eswa.2023.122406}

\bibitem{Ma2024UPCrackNetUP}
N.~Ma, R.~Fan, L.~Xie, \href{https://api.semanticscholar.org/CorpusID:267311537}{Up-cracknet: Unsupervised pixel-wise road crack detection via adversarial image restoration}, IEEE Transactions on Intelligent Transportation Systems 25 (2024) 13926--13936.
\newline\urlprefix\url{https://api.semanticscholar.org/CorpusID:267311537}

\bibitem{pmlr-v227-karimijafarbigloo24a}
S.~Karimijafarbigloo, R.~Azad, A.~Kazerouni, D.~Merhof, \href{https://proceedings.mlr.press/v227/karimijafarbigloo24a.html}{Ms-former: Multi-scale self-guided transformer for medical image segmentation}, in: I.~Oguz, J.~Noble, X.~Li, M.~Styner, C.~Baumgartner, M.~Rusu, T.~Heinmann, D.~Kontos, B.~Landman, B.~Dawant (Eds.), Medical Imaging with Deep Learning, Vol. 227 of Proceedings of Machine Learning Research, PMLR, 2024, pp. 680--694.
\newline\urlprefix\url{https://proceedings.mlr.press/v227/karimijafarbigloo24a.html}

\bibitem{CFD}
Y.~Shi, L.~Cui, Z.~Qi, F.~Meng, Z.~Chen, \href{https://ieeexplore.ieee.org/document/7471507}{Automatic road crack detection using random structured forests}, IEEE Transactions on Intelligent Transportation Systems 17~(12) (2016) 3434--3445.
\newline\urlprefix\url{https://ieeexplore.ieee.org/document/7471507}

\bibitem{CRACK500}
L.~Zhang, F.~Yang, Y.~D. Zhang, Y.~J. Zhu, \href{https://ieeexplore.ieee.org/document/7533052}{Road crack detection using deep convolutional neural network}, in: Image Processing (ICIP), 2016 IEEE International Conference on, IEEE, 2016, pp. 3708--3712.
\newline\urlprefix\url{https://ieeexplore.ieee.org/document/7533052}

\bibitem{cracktree200}
Q.~Zou, Y.~Cao, Q.~Li, Q.~Mao, S.~Wang, \href{https://doi.org/10.1016/j.patrec.2011.11.004}{Cracktree: Automatic crack detection from pavement images}, Pattern Recognition Letters 33~(3) (2012) 227--238.
\newline\urlprefix\url{https://doi.org/10.1016/j.patrec.2011.11.004}

\bibitem{eugen_miller}
S.~Ham, S.~Bae, H.~Kim, I.~Lee, G.-P. Lee, D.~Kim, \href{https://www.jkta.or.kr/articles/xml/kqgN/}{Training a semantic segmentation model for cracks in the concrete lining of tunnel}, Journal of Korean Tunnelling and Underground Space Association (2021) 549--558\href {https://doi.org/10.9711/KTAJ.2021.23.6.549} {\path{doi:10.9711/KTAJ.2021.23.6.549}}.
\newline\urlprefix\url{https://www.jkta.or.kr/articles/xml/kqgN/}

\bibitem{forest}
Y.~Shi, L.~Cui, Z.~Qi, F.~Meng, Z.~Chen, \href{https://ieeexplore.ieee.org/abstract/document/7471507}{Automatic road crack detection using random structured forests}, IEEE Transactions on Intelligent Transportation Systems 17~(12) (2016) 3434--3445.
\newline\urlprefix\url{https://ieeexplore.ieee.org/abstract/document/7471507}

\bibitem{GAPS384}
M.~Eisenbach, R.~Stricker, D.~Seichter, K.~Amende, K.~Debes, M.~Sesselmann, D.~Ebersbach, U.~Stoeckert, H.-M. Gross, \href{https://ieeexplore.ieee.org/document/7966101}{How to get pavement distress detection ready for deep learning? a systematic approach.}, in: International Joint Conference on Neural Networks (IJCNN), 2017, pp. 2039--2047.
\newline\urlprefix\url{https://ieeexplore.ieee.org/document/7966101}

\bibitem{Volker}
M.~Pak, S.~Kim, \href{https://doi.org/10.1007/978-981-15-9343-7_36}{Crack detection using fully convolutional network in wall-climbing robot}, in: J.~J. Park, S.~J. Fong, Y.~Pan, Y.~Sung (Eds.), Advances in Computer Science and Ubiquitous Computing, Springer Singapore, Singapore, 2021, pp. 267--272.
\newline\urlprefix\url{https://doi.org/10.1007/978-981-15-9343-7_36}

\bibitem{sylvie}
R.~Amhaz, S.~Chambon, J.~Idier, V.~Baltazart, \href{https://ieeexplore.ieee.org/document/7572082}{Automatic crack detection on two-dimensional pavement images: An algorithm based on minimal path selection}, IEEE Transactions on Intelligent Transportation Systems 17~(10) (2016) 2718--2729.
\newblock \href {https://doi.org/10.1109/TITS.2015.2477675} {\path{doi:10.1109/TITS.2015.2477675}}.
\newline\urlprefix\url{https://ieeexplore.ieee.org/document/7572082}

\bibitem{FCN}
E.~Shelhamer, J.~Long, T.~Darrell, \href{https://api.semanticscholar.org/CorpusID:1629541}{Fully convolutional networks for semantic segmentation}, 2015 IEEE Conference on Computer Vision and Pattern Recognition (CVPR) (2014) 3431--3440.
\newline\urlprefix\url{https://api.semanticscholar.org/CorpusID:1629541}

\bibitem{Unet}
O.~Ronneberger, P.~Fischer, T.~Brox, \href{https://link.springer.com/chapter/10.1007/978-3-319-24574-4_28}{U-net: Convolutional networks for biomedical image segmentation}, in: N.~Navab, J.~Hornegger, W.~M. Wells, A.~F. Frangi (Eds.), Medical Image Computing and Computer-Assisted Intervention -- MICCAI 2015, Springer International Publishing, Cham, 2015, pp. 234--241.
\newline\urlprefix\url{https://link.springer.com/chapter/10.1007/978-3-319-24574-4_28}

\bibitem{Unet++}
Z.~Zhou, M.~M.~R. Siddiquee, N.~Tajbakhsh, J.~Liang, \href{https://ieeexplore.ieee.org/document/8932614}{Unet++: Redesigning skip connections to exploit multiscale features in image segmentation}, IEEE Transactions on Medical Imaging 39~(6) (2020) 1856--1867.
\newblock \href {https://doi.org/10.1109/TMI.2019.2959609} {\path{doi:10.1109/TMI.2019.2959609}}.
\newline\urlprefix\url{https://ieeexplore.ieee.org/document/8932614}

\bibitem{PSPNet}
H.~Zhao, J.~Shi, X.~Qi, X.~Wang, J.~Jia, \href{https://ieeexplore.ieee.org/document/8100143}{Pyramid scene parsing network}, in: 2017 IEEE Conference on Computer Vision and Pattern Recognition (CVPR), 2017, pp. 6230--6239.
\newblock \href {https://doi.org/10.1109/CVPR.2017.660} {\path{doi:10.1109/CVPR.2017.660}}.
\newline\urlprefix\url{https://ieeexplore.ieee.org/document/8100143}

\bibitem{PAN}
H.~Li, P.~Xiong, J.~An, L.~Wang, \href{https://api.semanticscholar.org/CorpusID:44120007}{Pyramid attention network for semantic segmentation}, in: Proceedings of the British Machine Vision Conference (BMVC), 2018, p. 285.
\newblock \href {https://doi.org/10.48550/arXiv.1805.10180} {\path{doi:10.48550/arXiv.1805.10180}}.
\newline\urlprefix\url{https://api.semanticscholar.org/CorpusID:44120007}

\bibitem{MAnet}
R.~Li, S.~Zheng, C.~Zhang, C.~Duan, J.~Su, L.~Wang, P.~Atkinson, \href{https://api.semanticscholar.org/CorpusID:221507549}{Multiattention network for semantic segmentation of fine-resolution remote sensing images}, IEEE Transactions on Geoscience and Remote Sensing PP (2021) 1--13.
\newblock \href {https://doi.org/10.1109/TGRS.2021.3093977} {\path{doi:10.1109/TGRS.2021.3093977}}.
\newline\urlprefix\url{https://api.semanticscholar.org/CorpusID:221507549}

\bibitem{Linknet}
A.~Chaurasia, E.~Culurciello, \href{https://ieeexplore.ieee.org/document/8305148}{Linknet: Exploiting encoder representations for efficient semantic segmentation}, in: 2017 IEEE Visual Communications and Image Processing (VCIP), 2017, pp. 1--4.
\newblock \href {https://doi.org/10.1109/VCIP.2017.8305148} {\path{doi:10.1109/VCIP.2017.8305148}}.
\newline\urlprefix\url{https://ieeexplore.ieee.org/document/8305148}

\bibitem{FPN}
A.~Kirillov, R.~Girshick, K.~He, P.~Dollar, \href{https://ieeexplore.ieee.org/document/8954091}{Panoptic feature pyramid networks}, in: Proceedings of the IEEE/CVF Conference on Computer Vision and Pattern Recognition (CVPR), 2019, pp. 6392--6401.
\newblock \href {https://doi.org/10.1109/CVPR.2019.00656} {\path{doi:10.1109/CVPR.2019.00656}}.
\newline\urlprefix\url{https://ieeexplore.ieee.org/document/8954091}

\bibitem{chen2017rethinking}
L.-C. Chen, G.~Papandreou, F.~Schroff, H.~Adam, \href{https://api.semanticscholar.org/CorpusID:22655199}{Rethinking atrous convolution for semantic image segmentation}, in: Proceedings of the IEEE Conference on Computer Vision and Pattern Recognition (CVPR), 2017, pp. 1885--1894.
\newline\urlprefix\url{https://api.semanticscholar.org/CorpusID:22655199}

\bibitem{deeplabv3plus}
L.-C. Chen, Y.~Zhu, G.~Papandreou, F.~Schroff, H.~Adam, \href{https://link.springer.com/chapter/10.1007/978-3-030-01234-2_49}{Encoder-decoder with atrous separable convolution for semantic image segmentation}, in: V.~Ferrari, M.~Hebert, C.~Sminchisescu, Y.~Weiss (Eds.), Computer Vision -- ECCV 2018, Springer International Publishing, Cham, 2018, pp. 833--851.
\newline\urlprefix\url{https://link.springer.com/chapter/10.1007/978-3-030-01234-2_49}

\bibitem{UPerNet}
T.~Xiao, Y.~Liu, B.~Zhou, Y.~Jiang, J.~Sun, \href{https://api.semanticscholar.org/CorpusID:50781105}{Unified perceptual parsing for scene understanding}, in: European Conference on Computer Vision, 2018, pp. 432--448.
\newline\urlprefix\url{https://api.semanticscholar.org/CorpusID:50781105}

\bibitem{Segformer}
E.~Xie, W.~Wang, Z.~Yu, A.~Anandkumar, J.~M. {\'A}lvarez, P.~Luo, \href{https://api.semanticscholar.org/CorpusID:235254713}{Segformer: Simple and efficient design for semantic segmentation with transformers}, in: Neural Information Processing Systems, 2021, pp. 12077--12090.
\newline\urlprefix\url{https://api.semanticscholar.org/CorpusID:235254713}

\bibitem{RIAD}
V.~Zavrtanik, M.~Kristan, D.~Skočaj, \href{https://www.sciencedirect.com/science/article/pii/S0031320320305094}{Reconstruction by inpainting for visual anomaly detection}, Pattern Recognition 112 (2021) 107706.
\newblock \href {https://doi.org/https://doi.org/10.1016/j.patcog.2020.107706} {\path{doi:https://doi.org/10.1016/j.patcog.2020.107706}}.
\newline\urlprefix\url{https://www.sciencedirect.com/science/article/pii/S0031320320305094}

\bibitem{SCADN}
X.~Yan, H.~Zhang, X.~Xu, X.~Hu, P.-A. Heng, \href{https://ojs.aaai.org/index.php/AAAI/article/view/16420}{Learning semantic context from normal samples for unsupervised anomaly detection}, Proceedings of the AAAI Conference on Artificial Intelligence 35~(4) (2021) 3110--3118.
\newblock \href {https://doi.org/10.1609/aaai.v35i4.16420} {\path{doi:10.1609/aaai.v35i4.16420}}.
\newline\urlprefix\url{https://ojs.aaai.org/index.php/AAAI/article/view/16420}

\bibitem{UP-CrackNet}
N.~Ma, R.~Fan, L.~Xie, \href{https://dl.acm.org/doi/abs/10.1109/TITS.2024.3398037}{Up-cracknet: Unsupervised pixel-wise road crack detection via adversarial image restoration}, IEEE Transactions on Intelligent Transportation Systems PP (2024) 1--11.
\newblock \href {https://doi.org/10.1109/TITS.2024.3398037} {\path{doi:10.1109/TITS.2024.3398037}}.
\newline\urlprefix\url{https://dl.acm.org/doi/abs/10.1109/TITS.2024.3398037}

\bibitem{Mohammed}
M.~Mohammed, Z.~Han, Y.~Li, Z.~Al-Huda, C.~Li, W.-D. Wang, \href{https://www.frontiersin.org/journals/materials/articles/10.3389/fmats.2022.1058407/full}{End-to-end semi-supervised deep learning model for surface crack detection of infrastructures}, Frontiers in Materials 9 (12 2022).
\newblock \href {https://doi.org/10.3389/fmats.2022.1058407} {\path{doi:10.3389/fmats.2022.1058407}}.
\newline\urlprefix\url{https://www.frontiersin.org/journals/materials/articles/10.3389/fmats.2022.1058407/full}

\bibitem{Zhang2025DeepLF}
X.~Zhang, H.~Wang, Y.-A. Hsieh, Z.~Yang, A.~Yezzi, Y.-C. Tsai, \href{https://api.semanticscholar.org/CorpusID:280650235}{Deep learning for crack detection: A review of learning paradigms, generalizability, and datasets}, 2025.
\newline\urlprefix\url{https://api.semanticscholar.org/CorpusID:280650235}

\end{thebibliography}
\end{document}